\documentclass[runningheads]{llncs}

\usepackage{eccv}
\usepackage{eccvabbrv}

\usepackage{graphicx}
\usepackage{booktabs}

\usepackage[accsupp]{axessibility}  

\usepackage{hyperref}

\usepackage{orcidlink}
\usepackage{wrapfig}
\usepackage{colortbl}
\usepackage{makecell}
\usepackage{multirow,adjustbox,diagbox,threeparttable, tabularx}
\usepackage[normalem]{ulem}
\usepackage{caption}
\usepackage{microtype}

\usepackage[table,xcdraw]{xcolor}
\definecolor{Light}{rgb}{0.99, 0.92, 0.95}
\renewcommand{\paragraph}[1]{\vspace{1.25mm}\noindent\textbf{#1}}

\usepackage{booktabs}
\usepackage{amssymb} 

\definecolor{codegreen}{rgb}{0,0.6,0}
\definecolor{codegray}{rgb}{0.7,0.7,0.7}
\definecolor{codepurple}{rgb}{0.58,0,0.82}
\definecolor{backcolour}{rgb}{1.0,1.0,1.0}

\newcolumntype{C}{>{\centering\arraybackslash}X}

\newcommand{\method}{Loc3R-VLM}

\begin{document}

\title{Loc3R-VLM: Language-based Localization and 3D Reasoning with Vision-Language Models}

\titlerunning{Loc3R-VLM}

\author{Kevin Qu$^*$\inst{1,2} \and
Haozhe Qi\inst{3} \and
Mihai Dusmanu\inst{1} \and
Mahdi Rad\inst{1} \and
Rui Wang\inst{1} \and
Marc Pollefeys\inst{1,2}
}

\authorrunning{K.~Qu et al.}

\institute{$^1$Microsoft Spatial AI Lab \quad $^2$ETH Zurich \quad $^3$EPFL}

\maketitle

\let\thefootnote\relax\footnotetext{$^*$Work done during an internship at Microsoft}

\begin{center}
  \vspace{-1.2em}
  \includegraphics[width=0.95\textwidth]{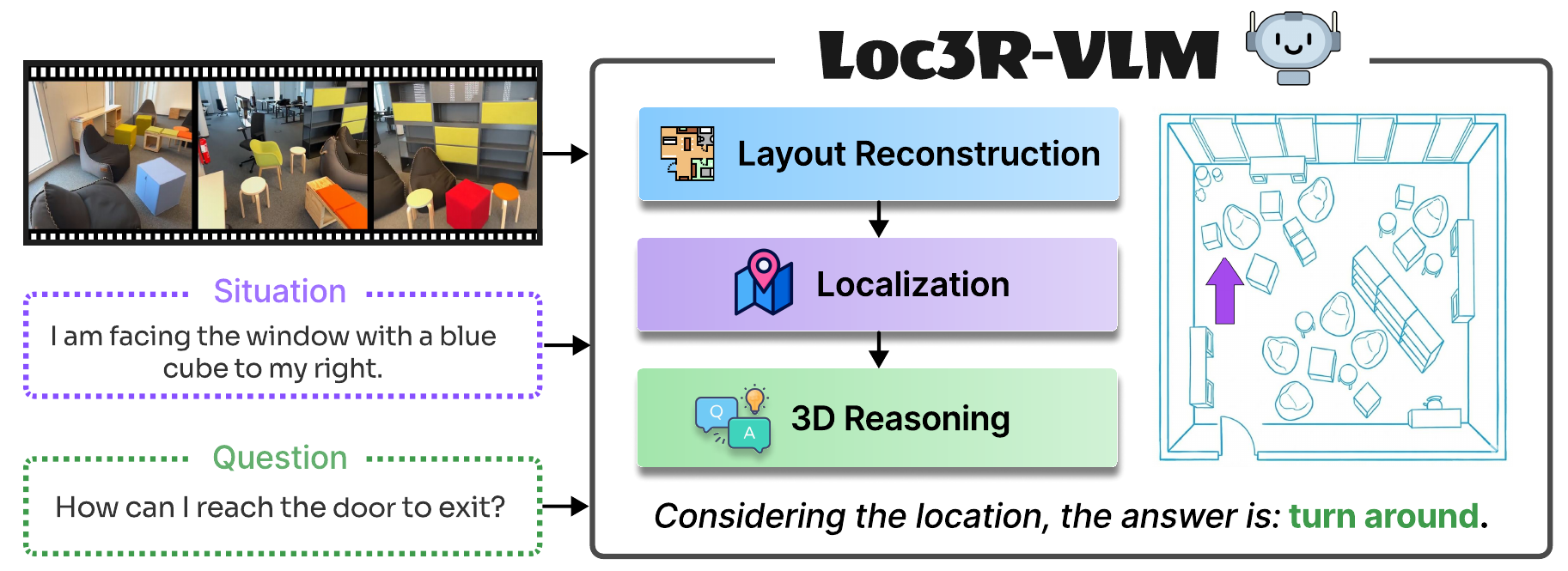}
  \captionof{figure}{
    \method{} equips 2D VLMs with advanced 3D spatial understanding capabilities from video. Inspired by human cognition, it builds an internal cognitive map of the global environment while explicitly modeling an agent's position and orientation. By jointly capturing global layout and egocentric state, the model excels at two core tasks: language-driven localization and viewpoint-aware 3D reasoning.}
  \label{fig:teaser}
\end{center}

\begin{abstract}
Multimodal Large Language Models (MLLMs) have made impressive progress in connecting vision and language, but they still struggle with spatial understanding and viewpoint-aware reasoning. Recent efforts aim to augment the input representations with geometric cues rather than explicitly teaching models to reason in 3D space. We introduce \method{}, a framework that equips 2D Vision–Language Models with advanced 3D understanding capabilities from monocular video input. 
Inspired by human spatial cognition, \method{} relies on two joint objectives: global layout reconstruction to build a holistic representation of the scene structure, and explicit situation modeling to anchor egocentric perspective. These objectives provide direct spatial supervision that grounds both perception and language in a 3D context. To ensure geometric consistency and metric-scale alignment, we leverage lightweight camera pose priors extracted from a pre-trained 3D foundation model. \method{} achieves state-of-the-art performance in language-based localization and outperforms existing 2D- and video-based approaches on situated and general 3D question-answering benchmarks, demonstrating that our spatial supervision framework enables strong 3D understanding. Project page: \href{https://kevinqu7.github.io/loc3r-vlm}{https://kevinqu7.github.io/loc3r-vlm}
\keywords{3D Scene Understanding \and Spatial Reasoning \and Language-based Localization \and Vision-Language Models}
\end{abstract}    
\section{Introduction}\label{sec:introduction} \label{sec:intro}

Humans naturally possess an intuitive grasp of their surroundings. When observing a scene, we construct a mental representation of the environment akin to a cognitive map that can be recalled and manipulated long after the initial perception~\cite{kosslyn1978,sep-mental-imagery,Paivio1979}. This ability enables us to answer spatial queries, such as locating objects or determining directions, by mentally repositioning ourselves within this map and imagining alternative viewpoints~\cite{Newcombe2024Spatial,Tolman1948,Bottini2020KnowledgeAR}.
Replicating such visual–spatial intelligence in artificial systems remains a significant challenge~\cite{yang2024thinkinginspace,lee2025perspective}. Although Multimodal Large Language Models (MLLMs) have made rapid progress in linking language with 2D imagery~\cite{zhang2024llava,chen2024internvl2,Qwen2-VL,comanici2025gemini25pushingfrontier,openai2024gpt4technicalreport}, they still lack a coherent understanding of 3D space~\cite{zhang2025mllmsstrugglespatialunderstanding,yang2024thinkinginspace,kamath2023whatsup,chen2025spatialreasoninghardvlms}. Most MLLMs operate in a local manner, struggling to integrate observations across multiple frames into a persistent, unified global context~\cite{xu2025multi}. 

Recent research has increasingly focused on enhancing the spatial awareness of MLLMs. Two common approaches have emerged: (i) encoding point cloud representations directly into the model~\cite{zhang2024chatscene,deng20253dllava,zhu20233dvista, yu2025inst3dllm,huang2024chatscene}, and (ii) augmenting 2D image inputs with 3D positional encodings derived from depth maps and camera poses~\cite{zheng2024video3dllm,zhu2024llava3d, cheng20253dawareregionprompted}. 
However, both strategies suffer from two fundamental limitations. First, these approaches often require precise 3D ground-truth data during inference, which is rarely available in real-world settings. Second, even when 3D-augmented inputs are provided, their supervision typically only focuses on language-based or object-centric objectives. 
Since global scene understanding and situational awareness are treated as mere byproducts rather than explicitly learned capabilities, these models frequently fail to reason about viewpoint-dependent spatial relationships or infer perspectives beyond the camera’s egocentric view
~\cite{lee2025perspective,goral2024seeingeyesevaluatingvisual,zhang2025do, zhang2025sphere}. These shortcomings are particularly critical in domains such as robotics~\cite{open_x_embodiment_rt_x_2023,driess2023palme,procthor,rt22023arxiv} or autonomous driving~\cite{tian2024DriveVLM,Kong_2025_vlrdriver,ma2024dolphins}, where situational understanding underpins safe navigation and decision-making. Despite its importance, explicit situation modeling remains relatively underexplored and is largely confined to point-cloud–based methods~\cite{ma2022sqa3d,yuan2025empoweringsituation,man2024sig3d,zhu20233dvista}, which face scalability and generalization challenges due to the scarcity of paired 3D–text data.

To overcome these limitations, we introduce \method{}, a novel framework that endows 2D Vision-Language Models (VLMs) with advanced 3D reasoning capabilities and situational awareness. We draw inspiration from human cognition by focusing on two key capabilities:
(1) Inspired by how humans form a cognitive map of a scene, \method{} learns to reconstruct the global layout, enabling the model to maintain an internal memory of the environment and capture its spatial organization. (2) Mirroring our ability to imagine any viewpoint within a space, \method{} incorporates explicit situation modeling, allowing the model to localize itself within the scene and reason from that grounded perspective. To reinforce geometric consistency, we integrate a camera pose prior from a pre-trained 3D foundation model, ensuring alignment in pose and scale. By unifying these components within a joint training framework, \method{} bridges the gap between visual perception, spatial understanding, and embodied reasoning.

\method{} achieves state-of-the-art performance in language-based localization and surpasses existing video-based models on both general and situated 3D question-answering benchmarks. This work underscores the importance of explicit spatial supervision and situational awareness, bringing us closer to models capable of perceiving and reasoning about the world with human-like spatial understanding.
To summarize, our main contributions include:
\begin{itemize}
\item We propose Loc3R-VLM, a framework that equips a 2D Vision-Language Model with advanced 3D understanding capabilities from monocular video input.
\item We introduce a 3D-aware learning strategy that combines a global layout reconstruction objective for holistic scene understanding with an explicit situation modeling module for localization and perspective-aware reasoning.
\item We develop a lightweight mechanism that integrates a camera pose prior from a pre-trained 3D foundation model for stable geometric grounding.
\item \method{} significantly outperforms existing methods in language-based localization and surpasses video-based approaches on both situated and general 3D question-answering benchmarks.
\end{itemize}

\section{Related Work}\label{sec:related_work}
\subsection{MLLMs for 3D Scene Understanding}\label{sec:mllms_for_scene}
Recent advances in Multimodal Large Language Models aim to extend their reasoning capabilities from text and 2D to the 3D domain. Earlier approaches~\cite{chen2023ll3da,zhu20pq3d,hong20233dllm,huang2024leo,zhang2024chatscene,deng20253dllava,zhu20233dvista,huang2023chat3dv2,fu2024scenellmextendinglanguagemodel,yu2025inst3dllm,zhi2024lscenellm,huang2024chatscene,kang2024robin3d} adopt point clouds as the underlying scene representation and propose strategies for extracting geometric and semantic features before aligning them with the language space of the LLM. However, the scarcity of large-scale paired 3D–text data remains a major bottleneck for generalization.

To overcome these constraints, recent work shifts focus from 3D point cloud MLLMs to leveraging multi-view image or video inputs, exploiting the strong 2D priors of pre-trained Vision-Language Models (VLMs)~\cite{zhang2024llava,chen2024internvl2,Qwen2-VL,comanici2025gemini25pushingfrontier,openai2024gpt4technicalreport}. LLaVA-3D~\cite{zhu2024llava3d} and Video3D-LLM~\cite{zheng2024video3dllm} incorporate 3D positional information by augmenting 2D patch features with 3D coordinate embeddings. Ross3D~\cite{wang2025ross3d} further extends Video3D-LLM with reconstructive visual instruction tuning, providing 3D-aware supervision through cross-view and global reconstruction tasks. While conceptually aligned with our goal of learning a global scene representation, Ross3D and related approaches require accurate ground-truth camera poses and depth maps to compute the 3D coordinate embeddings — inputs that are rarely available for unconstrained video data.

Most recently, researchers have begun leveraging internal representations from 3D foundation models~\cite{wang2025cut3r,wang2025vggt} to provide implicit geometric cues~\cite{zheng2025learningvideos3dworld,fan2025vlm3rvisionlanguagemodelsaugmented,wu2025spatialmllmboostingmllmcapabilities} to VLMs. While promising, these methods typically use this spatial information as a mere input augmentation or additional feature stream, rather than explicitly teaching the model 3D awareness.
In contrast, our framework moves beyond passive input augmentation. By training with explicit spatial supervision, we enable robust spatial understanding directly from monocular videos, eliminating the reliance on ground-truth 3D annotations during inference.

\subsection{Language-based Localization}
Language-based localization research explores two distinct directions. The first line of work tackles text-to-3D localization in large-scale outdoor environments ~\cite{xia2024text2loc,wang2023ret,xu2025cmmlocadvancingtexttopointcloudlocalization,wang2024instancefreetextpointcloud}. These approaches are designed for outdoor LiDAR data and typically support only coarse spatial grounding, with limited open-set language generalization and a lack of orientation estimation.

We focus on language-based localization in indoor scenes, where viewpoint ambiguity, occlusions, and fine-grained object relationships pose unique challenges. This task requires inferring an agent’s position and orientation directly from a natural language situation description. Prior works in this domain represent scenes using dense 3D geometries, such as point clouds or voxel grids ~\cite{man2024sig3d,yuan2025empoweringsituation,zhu20233dvista,ma2022sqa3d}.
SQA3D~\cite{ma2022sqa3d} fuses textual inputs with object-level 3D features through cross-attention and employs auxiliary heads to predict position and orientation. SIG3D~\cite{man2024sig3d} proposes a situation-grounded pipeline that voxelizes the scene and performs anchor-based prediction of position and rotation. The estimated pose is used to re-encode visual tokens to enable downstream viewpoint-aware reasoning. View2Cap~\cite{yuan2025empoweringsituation} proposes a situation grounding module. It encodes object point cloud instances as visual tokens and classifies offsets and orientation bins relative to anchor objects to recover the final pose. 

A fundamental limitation of these existing methods is their reliance on dense point-cloud representations, which severely restricts scalability and hinders generalization. 
In contrast, our approach operates directly on monocular video, enabling practical inference from easily accessible visual data.

\begin{figure*}[!t]
    \centering
    \includegraphics[width=\linewidth]{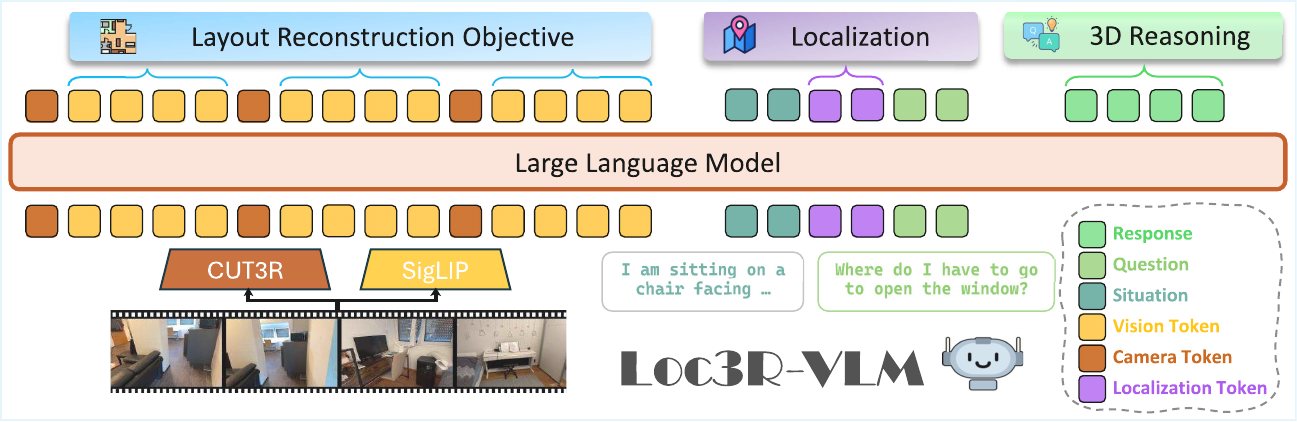}
    \vspace{-0.55cm}
    \caption{
    \textbf{Overview of \method{}}.  
    Our framework takes a monocular video as input and augments the vision token sequence with per-frame latent camera pose priors extracted from the 3D foundation model CUT3R~\cite{wang2025cut3r}. The model is jointly trained using two spatial objectives: (1) layout reconstruction, which grounds vision patch tokens into a bird's-eye-view (BEV) space to capture global scene structure, and (2) situation modeling, which utilizes dedicated localization query tokens to localize an agent from a situation description. During answer generation, the model leverages the inferred layout and location to perform viewpoint-aware 3D reasoning.
    }
    \label{fig:pipeline}
\end{figure*}

\vspace{-0.1cm}
\section{Method}\label{sec:method}
\method{} equips a VLM with 3D spatial understanding and situational awareness capabilities directly from monocular video input. An overview of our method is illustrated in~\cref{fig:pipeline}. Our framework consists of three complementary components.
First, we incorporate lightweight Camera Pose Priors (\cref{sec:cam_token}), where latent embeddings from a pre-trained 3D foundation model supply geometric cues that mitigate the inherent scale ambiguity of monocular video and support the VLM for localization in metric space.
Building on these priors, our Global Layout Reconstruction component (\cref{sec:layout_rec}) encourages the model to form a coherent bird’s-eye-view (BEV) representation of the scene. This enables capturing object placements, cross-view spatial relationships, and global context into a unified representation.
To enable situational awareness, we further introduce a Situation Modeling mechanism (\cref{sec:situation_modeling}) that explicitly represents the agent’s position and orientation, allowing the model to perform localization and viewpoint-aware inference from natural language descriptions.
Finally, \cref{sec:overall_loss} presents the unified training objective that jointly optimizes these components within a single multimodal framework.

\vspace{-0.1cm}
\subsection{Integration of Camera Pose Priors}\label{sec:cam_token}  
To spatially ground the input video, we incorporate per-frame latent camera tokens extracted from the pre-trained feed-forward geometry model CUT3R~\cite{wang2025cut3r}.
For each video frame $I_t$, CUT3R encodes the image through a vision transformer to produce feature tokens $F_t = f_{\text{enc}}(I_t)$. 
A learnable camera query token $z$ is prepended and processed with the previous recurrent state $s_{t-1}$ by a transformer decoder:
\begin{equation}
[z'_t, F'_t],\, s_t = f_{\text{dec}}([z, F_t], s_{t-1})
\end{equation}
The resulting camera token $z'_t$ and geometry tokens $F'_t$ jointly capture the current observation along with accumulated scene context, from which camera transformations and metric-scale point maps can be derived. 
Unlike prior works that fuse both camera and geometry tokens via cross-attention~\cite{fan2025vlm3rvisionlanguagemodelsaugmented} or inject geometry tokens via addition~\cite{wu2025spatialmllmboostingmllmcapabilities,zheng2025learningvideos3dworld}, we exclusively prepend the camera token to the vision token sequence for each frame. This strategy provides a stable geometric anchor that encodes pose priors while preserving the integrity of the pre-trained vision-language feature space of the VLM.

To integrate these pose priors into the VLM, we project the camera token $z'_t$ into the language embedding space using a learnable projection layer, implemented as two-layer MLP $f_{\text{cam}}$, yielding $c_t = f_{\text{cam}}(z'_t)$.  
Then, for each frame, the projected camera token is prepended to the vision tokens obtained by the SigLIP~\cite{zhai2023sigmoid} encoder to form the augmented vision sequence:
\begin{equation}
X_t^{aug} = [\,c_t,\, v_{t,1},\, v_{t,2},\, \ldots,\, v_{t,n}\,]
\end{equation}
This formulation embeds latent metric pose information directly into the visual stream, grounding every frame within the broader scene context.

\vspace{-0.1cm}
\subsection{Global Layout Reconstruction}\label{sec:layout_rec}
\begin{figure}[t]
    \centering
    \begin{minipage}[t]{0.49\linewidth}
        \vspace{0.1cm}
        \includegraphics[width=\linewidth]{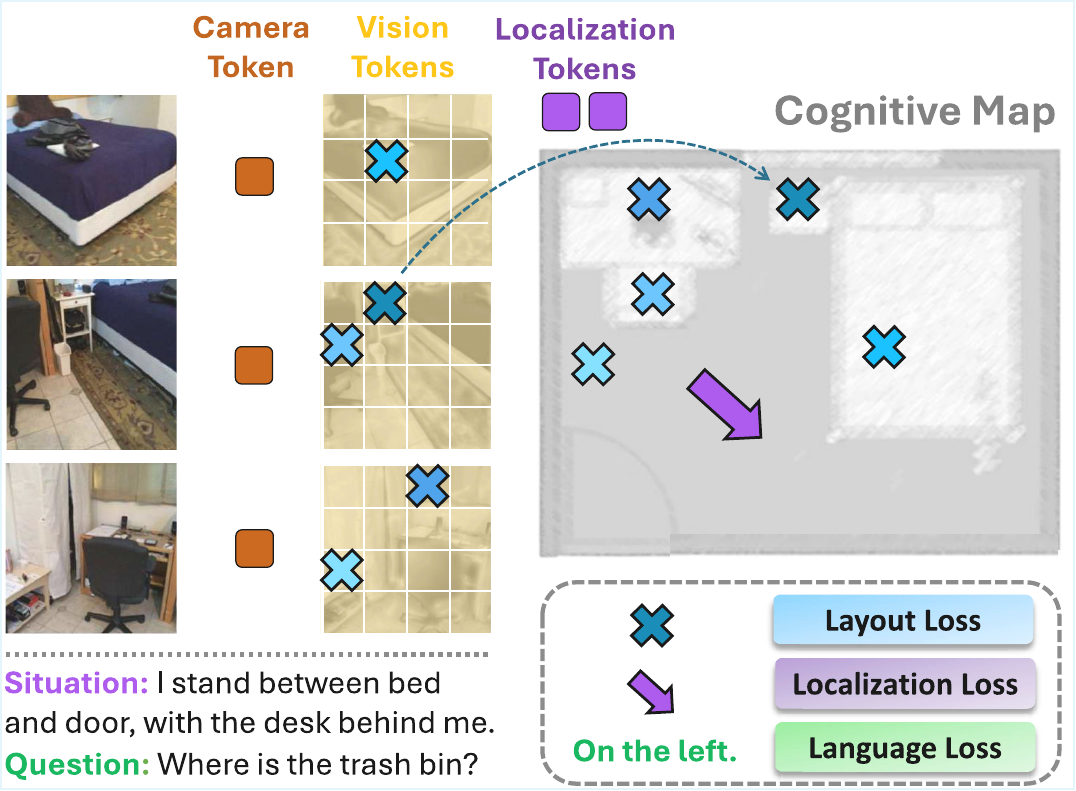}
    \end{minipage}\hfill
    \begin{minipage}[t]{0.48\linewidth}
        \caption{\textbf{Spatial Supervision Framework} introduces complementary training signals. For the layout reconstruction objective, the model learns to ground each vision patch token onto its corresponding BEV coordinate in a cognitive map to capture global scene structure. For localization, dedicated localization tokens explicitly model the agent's position and orientation. The framework is trained end-to-end using a joint objective of layout, localization, and language losses.}
        \label{fig:method}
    \end{minipage}
\end{figure}

Global Layout Reconstruction serves as an auxiliary training objective that enhances the model’s understanding of cross-view spatial relationships and global scene structure. In this task, the model learns to associate vision patch tokens with their corresponding two-dimensional coordinates within a unified bird’s-eye-view (BEV) representation, as illustrated in \cref{fig:method}. Importantly, this enables the model to ground observations across multiple frames into a persistent global context. This design is inspired by the way humans form cognitive maps of their environment, organizing spatial information into a lower-dimensional abstraction~\cite{Bottini2020KnowledgeAR, Tolman1948}.

The BEV space is defined in a gravity-aligned world coordinate frame that is shared consistently across all camera views from the same video. 
Following the coordinate system convention of CUT3R\cite{wang2025cut3r}, we anchor the world frame to the first video frame.
Given a sequence of \(M\) vision tokens \((\mathbf{v}_i)_{i=1}^{M}\) from the output layer of the LLM, we apply a learnable projection head \(f_{\text{proj}}\) to estimate each token’s spatial location in the BEV plane, alongside its associated predictive uncertainty:
\begin{equation}
[\hat{\mathbf{p}}_i,\, \boldsymbol{\hat{\sigma}}_i] = f_{\text{proj}}(\mathbf{v}_i)
\end{equation}
where \(\hat{\mathbf{p}}_i = [\hat{x}_i, \hat{y}_i]^\top \in \mathbb{R}^2\) denotes the predicted BEV position
and \(\boldsymbol{\hat{\sigma}}_i = [\hat{\sigma}_{x,i}, \hat{\sigma}_{y,i}]^\top \in \mathbb{R}^2\) represents the estimated uncertainty along each axis.

We model the ground-truth BEV coordinates \(\mathbf{p}_i = [x_i, y_i]^\top\)
as a sample drawn from a Gaussian distribution centered at \(\hat{\mathbf{p}}_i\)
with diagonal covariance matrix \(\mathrm{diag}(\hat{\sigma}_{x,i}^2,\, \hat{\sigma}_{y,i}^2)\).
The training objective minimizes the Gaussian negative log-likelihood~\cite{kendall2017uncertainties}:
\begin{align}
\mathcal{L}_{\text{BEV}}
= \frac{1}{M}\!\sum_{i=1}^{M} \frac{1}{2} \bigg[ &
\frac{(x_i-\hat{x}_i)^2}{\hat{\sigma}_{x,i}^2} + \log(\hat{\sigma}_{x,i}^2) + \nonumber
\frac{(y_i-\hat{y}_i)^2}{\hat{\sigma}_{y,i}^2}
+ \log(\hat{\sigma}_{y,i}^2) \bigg] \enspace
\end{align}
This loss encourages the model to build a coherent global representation of the scene, while enriching the hidden states of the vision tokens with spatial information. Further details on the BEV representation are provided in the Supplementary Material.

\subsection{Situation Modeling}\label{sec:situation_modeling}
To enable explicit localization and situation-aware reasoning, we introduce two new tokens \texttt{<Pos>} and \texttt{<Ori>} to the vocabulary, representing position and orientation, respectively. 
Given a textual situation description $\texttt{txt}_{\text{sit}}$ and a corresponding question $\texttt{txt}_{\text{q}}$, these tokens are inserted between the two text segments before tokenization:
\begin{equation}
X_{\text{in}} = \texttt{concat}(\texttt{txt}_{\text{sit}},\, \texttt{<Pos>},\, \texttt{<Ori>},\, \texttt{txt}_{\text{q}}) \enspace 
\end{equation}

At the output layer of the LLM, the final hidden state of each token is decoded through lightweight task-specific heads. The position head estimates the agent’s 2D location $\hat{\mathbf{p}} = [\hat{x}, \hat{y}]^\top$ in the global BEV frame (Sec.~\ref{sec:layout_rec}), while the orientation head predicts discretized angle logits $\hat{\mathbf{y}}_{\text{ori}} \in \mathbb{R}^B$ over $B$ bins:
\begin{equation}
[\hat{\mathbf{p}},\, \boldsymbol{\sigma}_{\text{pos}}] = f_{\text{pos}}(\texttt{<Pos>})
\qquad
\hat{\mathbf{y}}_{\text{ori}} = f_{\text{ori}}(\texttt{<Ori>})
\end{equation}
Given that these new localization tokens are placed after the tokenized video input sequence, they can causally attend to both the camera tokens from~\cref{sec:cam_token} and the spatially-enriched vision tokens from~\cref{sec:layout_rec}.

\paragraph{Position Estimation.}
The position head predicts both position $\hat{\mathbf{p}}$ and its uncertainty $\boldsymbol{\sigma}_{\text{pos}} = [\sigma_x, \sigma_y]^\top$ in the same coordinate system defined in Sec.~\ref{sec:layout_rec}. 
We use the Gaussian negative log-likelihood (GNLL) loss defined in Sec.~\ref{sec:layout_rec} to supervise the predicted position. 
This probabilistic formulation not only down-weights ambiguous samples during training via the GNLL loss, but also teaches the model to output higher uncertainty for difficult cases, allowing the question-answering component to properly account for unreliable position estimates.

\paragraph{Orientation Estimation.}
The orientation angle $\theta \in [-\pi, \pi)$ is discretized into $B$ uniform bins with centers $\{\theta_b\}_{b=1}^{B}$. 
To provide a smooth training signal, we construct a wrapped Gaussian target distribution centered at the ground-truth angle:
\begin{equation}
w_b =
\exp\!\left(
-\tfrac{1}{2}
\left[
\frac{\mathrm{wrap}(\theta - \theta_b)}{\sigma_{\text{ori}}}
\right]^2
\right)
\end{equation}
where $\mathrm{wrap}(\cdot)$ maps angular differences into $[-\pi, \pi)$. 
We then normalize these weights across bins to obtain a valid probability distribution
\begin{equation}
y_{\text{ori}}^{(b)} = \frac{w_b}{\sum_{b'=1}^{B} w_{b'}}
\end{equation}
The orientation head outputs logits $\hat{\mathbf{y}}_{\text{ori}} \in \mathbb{R}^B$, which are supervised using a KL-divergence loss:
\begin{equation}
\mathcal{L}_{\text{ori}}
=
\mathrm{KL}\!\left(
\mathbf{y}_{\text{ori}} \,\|\, \mathrm{softmax}(\hat{\mathbf{y}}_{\text{ori}})
\right)
\end{equation}
This circular formulation ensures stable gradients near angle boundaries and avoids discontinuities. 

At inference, we recover a continuous orientation estimation using a circular soft-argmax.
Let $p_b = \mathrm{softmax}(\hat{\mathbf{y}}_{\text{ori}})_b$ denote the predicted probabilities. 
We first compute the expectation on the unit circle:
\begin{equation}
\mathbf{\hat{v}} = \sum_{b=1}^{B} p_b
\begin{bmatrix}
\cos\theta_b \\ \sin\theta_b
\end{bmatrix}
\end{equation}
and recover the final angle as:
\begin{equation}
\hat{\theta} = \mathrm{atan2}(\hat{v}_y, \hat{v}_x).
\end{equation}

\paragraph{Joint Situation Objective.}
The final situation modeling objective combines both components:
\begin{equation}
\mathcal{L}_{\text{sit}} = \mathcal{L}_{\text{pos}} + \lambda_{\text{ori}}\,\mathcal{L}_{\text{ori}} \enspace
\end{equation}
where we set the weighting coefficient $\lambda_{\text{ori}}=3.5$ to balance the magnitudes of the two loss terms. By introducing the explicit situation estimation objective, the model learns to represent and reason about the agent’s egocentric situation. 
Importantly, the explicit \texttt{<Pos>} and \texttt{<Ori>} tokens provide a dedicated representation for the agent’s situational state. 
During answer generation, the model can attend to these tokens to perform internal viewpoint transformation needed for situation-aware reasoning.

\subsection{Overall Training Objective}\label{sec:overall_loss}
\method{} is trained end-to-end with a joint objective that combines standard language modeling with our proposed spatial objectives. This allows the model to share a single multimodal representation across language, reconstruction, and situation objectives. The total loss is given by
\begin{equation}
\mathcal{L}_{\text{total}}
= \mathcal{L}_{\text{CE}}
+ \lambda_{\text{BEV}}\, \mathcal{L}_{\text{BEV}}
+ \lambda_{\text{sit}}\, \mathcal{L}_{\text{sit}} \enspace
\end{equation}
where $\mathcal{L}_{\text{CE}}$ denotes the autoregressive cross-entropy language modeling loss:
\begin{equation}
\mathcal{L}_{\text{CE}}
= - \frac{1}{T} \sum_{t=1}^{T} \log P_\theta (y_t \mid y_{<t}, X)
\end{equation}
with $X$ representing the input context and $y_{1:T}$ the target answer text tokens. We set the weighting coefficients to $\lambda_{\text{BEV}}=0.05$ and $\lambda_{\text{sit}}=0.075$ to balance between language and spatial loss contributions.

\begin{figure}[t]
    \centering
    \includegraphics[width=\linewidth]{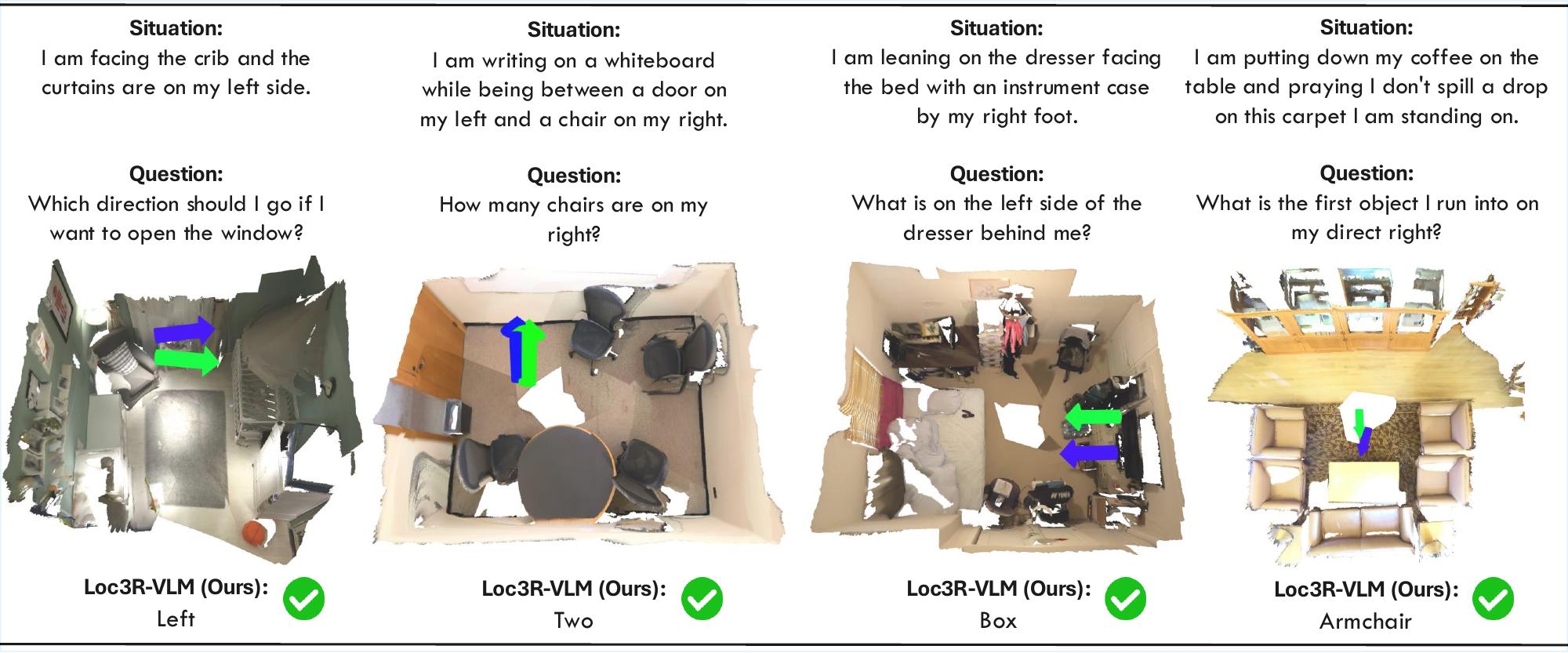}
    \vspace{-0.55cm}
    \caption{
    \textbf{Qualitative Results} for language-based localization and situated QA on SQA3D~\cite{ma2022sqa3d}. \method{} accurately grounds the described situations (\textcolor{blue}{blue}: prediction, \textcolor{green}{green}: ground truth) and provides the correct viewpoint-dependent answer. Meshes are shown for visualization only and are not used by the model.
    }
    \label{fig:qualitative}
\end{figure}

\section{Experiments}\label{sec:experiments}
\subsection{Implementation}\label{sec:implementation}
 We build \method{} on LLaVA-Video-7B~\cite{zhang2024llava} and fine-tune it using the training splits of ScanQA~\cite{azuma2022scanqa}, SQA3D~\cite{ma2022sqa3d}, the ScanNet~\cite{dai2017scannet} portion of MSQA~\cite{linghu2024msr3d}, and VSI-Bench~\cite{yang2024thinkinginspace} (official training split and custom training data curated by~\cite{fan2025vlm3rvisionlanguagemodelsaugmented}). Detailed dataset statistics are provided in the Supplementary Material. 
Training is performed for one epoch (4.2k steps) with a global batch size of 64 using AdamW optimizer and a cosine learning-rate schedule peaking at $1\times10^{-5}$. We update the parameters of the LLM, spatial and situation heads, and projection layers, while keeping the vision and CUT3R encoders frozen.
The spatial head is implemented as a single linear layer, and both situation heads use two-layer MLPs. The number of orientation bins is $B=36$ and the standard deviation is set to $\sigma_{ori}=2$.
All experiments are conducted on 16 NVIDIA Tesla V100 GPUs. Each scene is represented by 32 uniformly sampled video frames at an input resolution of $384\times384$. For question samples without a corresponding situation description, we create the pseudo-situation \textit{``I am standing in the room''}. During training, we utilize the depth and camera poses provided by the datasets to compute the ground-truth BEV coordinates for supervision of the layout reconstruction objective. At inference time, the model requires only raw monocular video as input, and no 3D annotations are needed.

\subsection{Evaluation}\label{sec:evaluation}
We assess the spatial understanding capabilities of \method{} on language-based localization, situated 3D question answering, and general 3D question answering tasks across several benchmarks. Qualitative examples of our method are illustrated in~\cref{fig:qualitative}.

\paragraph{Language-based Localization Benchmarks.}
We evaluate localization performance on SQA3D~\cite{ma2022sqa3d}, which contains text scenarios describing an embodied agent’s situation in the scene. The test split includes 719 samples across 67 indoor scenes sourced from ScanNet~\cite{dai2017scannet}.

\begin{table}[t]
\centering
\caption{\textbf{Evaluation of Language-based Localization} on SQA3D~\cite{ma2022sqa3d}. $^\dagger$ indicates results reproduced by~\cite{yuan2025empoweringsituation}. \textit{Separate} means disabling other tasks to let the model focus on situation estimation only. \method{} achieves state-of-the-art performance by a large margin and outperforms methods which operate on dense 3D point cloud input.}
\vspace{-0.2cm}
\resizebox{\textwidth}{!}{
\begin{tabular*}{\linewidth}{@{\extracolsep{\fill}} lcccccc @{}}
\toprule
\multirow{2}{*}{\textbf{Method}} & 
\multirow{2}{*}{\textbf{3D}} & 
\multirow{2}{*}{\textbf{2D}} &
\multicolumn{2}{c}{\textbf{Localization}} & 
\multicolumn{2}{c}{\textbf{Orientation}} \\
\cmidrule(lr){4-5} \cmidrule(lr){6-7}
 &  &  & Acc@0.5m & Acc@1.0m & Acc@15$^\circ$ & Acc@30$^\circ$ \\
\midrule
Random & -- & -- & 7.2 & 25.8 & 8.4 & 16.9 \\
SQA3D~\cite{ma2022sqa3d} & \checkmark & -- & 9.5 & 29.6 & 8.7 & 16.5 \\
SQA3D~\cite{ma2022sqa3d} \textit{(separate)} & \checkmark & -- & 10.3 & 31.4 & 17.1 & 22.8 \\
3D-VisTA~\cite{zhu20233dvista} & \checkmark & -- & 11.7 & 34.5 & 16.9 & 24.2 \\
SIG3D$^\dagger$~\cite{man2024sig3d} & \checkmark & -- & 16.8 & 35.2 & 23.4 & 26.3 \\
\textcolor{gray}{SIG3D}~\cite{man2024sig3d} & \checkmark & -- & \textcolor{gray}{27.4} & \textcolor{gray}{59.1} & \textcolor{gray}{28.7} & \textcolor{gray}{42.5} \\
View2Cap~\cite{yuan2025empoweringsituation} & \checkmark & -- & 17.4 & 36.9 & 24.1 & 28.5 \\
\midrule
Ours & -- & \checkmark & \textbf{42.6} & \textbf{75.9} & \textbf{38.4} & \textbf{63.0} \\
\bottomrule
\end{tabular*}

}
\label{table:localization}
\end{table}

\paragraph{Evaluation Metrics.}
We follow standard protocol~\cite{ma2022sqa3d,yuan2025empoweringsituation,man2024sig3d} and report both position and orientation accuracy. Acc@0.5m and Acc@1.0m measure the percentage of position predictions within 0.5\,m and 1.0\,m of the ground truth on the x-y plane, while Acc@15° and Acc@30° indicate the percentage of orientation predictions within 15° and 30° of the ground-truth yaw rotation.
\method{} predicts agent poses in a self-consistent, ego-centric BEV coordinate frame anchored to the first video frame. To compare these predictions with the dataset ground truth, which is defined in an arbitrary global coordinate system, we use the camera pose of the first frame to rigidly align the BEV frame to this global coordinate system for evaluation. This alignment involves only rotation and translation (no scaling) and is applied solely for computing the above metrics; the model does not require any ground-truth 3D annotations during inference.

\paragraph{Comparison with other methods.}
Table~\ref{table:localization} compares \method{} with existing language-based localization approaches. These include SQA3D~\cite{ma2022sqa3d}, 3D-VisTA~\cite{zhu20233dvista}, SIG3D~\cite{man2024sig3d}, and View2Cap~\cite{yuan2025empoweringsituation}, all of which rely on dense point-cloud representations as input. Despite not using explicit 3D input data, \method{} significantly outperforms all prior methods. In particular, it surpasses the strongest baseline View2Cap by $+25.2\%$ (Acc@0.5m) and $+39.0\%$ (Acc@1.0m) for position estimation, and by $+14.3\%$ (Acc@15$^\circ$) and $+34.5\%$ (Acc@30$^\circ$) for orientation prediction. These results highlight the effectiveness of our spatial supervision framework. The global layout reconstruction objective equips the model with a coherent mental map of the scene, while the camera pose prior provides metric-scale pose cues that stabilize absolute position estimation. Building on this internal implicit map, the situation modeling module enables robust viewpoint understanding, driving the large observed gains over the baselines.

\begin{table}[t]
    \scriptsize
    \centering
    \caption{\textbf{Evaluation of 3D QA} on VSI-Bench~\cite{yang2024thinkinginspace}. \method{} sets a new state-of-the-art and particularly excels at viewpoint-dependent subcategories.} 
    \vspace{-0.2cm}
    \resizebox{\linewidth}{!}{
    \begin{tabular*}{\linewidth}{@{\extracolsep{\fill}} l c c c c c c c c c @{}}
\toprule
\multirow{2}{*}{\textbf{Model}} 
& \multirow{2}{*}{\textbf{Avg.}} 
& \multicolumn{4}{c}{\textbf{Numerical}} 
& \multicolumn{4}{c}{\textbf{Multiple-Choice}} \\
\cmidrule(lr){3-6} \cmidrule(lr){7-10}
& & 
\makecell{Obj.\\Count} &
\makecell{Abs.\\Dist.} &
\makecell{Obj.\\Size} &
\makecell{Room\\Size} &
\makecell{Rel.\\Dist.} &
\makecell{Rel.\\Dir.} &
\makecell{Route\\Plan} &
\makecell{Appr.\\Order} \\
\midrule

\rowcolor{Light}\multicolumn{10}{l}{\textit{Expert Models}} \\
VLM-3R~\cite{fan2025vlm3rvisionlanguagemodelsaugmented}  & 60.9 & 70.2 & 49.4 & 69.2 & 67.1 & 65.4 & 80.5 & 45.4 & 40.1 \\
\midrule
\rowcolor{Light}\multicolumn{10}{l}{\textit{2D MLLMs}} \\
GPT-4o~\cite{openai2024gpt4technicalreport} & 34.0 & 46.2 & 5.3 & 43.8 & 38.2 & 37.0 & 41.3 & 31.5 & 28.5 \\
Gemini-1.5-Pro~\cite{comanici2025gemini25pushingfrontier} & 45.4 & 56.2 & 30.9 & 64.1 & 43.6 & 51.3 & 46.3 & 36.0 & 34.6 \\
InternVL2-8B~\cite{chen2024internvl2} & 34.6 & 23.1 & 28.7 & 48.2 & 39.8 & 36.7 & 30.7 & 29.9 & 39.6 \\
Qwen2.5-VL-7B~\cite{Qwen2-VL} & 33.0 & 40.9 & 14.8 & 43.4 & 10.7 & 38.6 & 38.5 & 33.0 & 29.8 \\
SAT-LLaVA-Video-7B~\cite{ray2025satdynamicspatialaptitude} & -- & -- & -- & -- & 47.3 & 41.1 & 37.1 & 36.1 & 40.4 \\
SPAR-8B~\cite{zhang2025spar} & 41.1 & -- & -- & -- & -- & -- & -- & -- & -- \\
ViLaSR-7B~\cite{wu2025reinforcingspatialreasoningvisionlanguage} & 45.4 & 63.5 & 34.4 & 60.6 & 30.9 & 48.9 & 45.2 & 30.4 & 49.2 \\
SpatialMLLM-4B~\cite{wu2025spatialmllmboostingmllmcapabilities} 
& 48.4 & 65.3 & 34.8 & 63.1 & 45.1 & 41.3 & 46.2 & 33.5 & 46.3 \\
Struct2D (7B)~\cite{zhu2025struct2d} 
& 41.9 & 46.0 & 34.7 & 56.4 & 42.6 & 35.1 & 44.9 & 33.5 & -- \\
VG-LLM-8B~\cite{zheng2025learningvideos3dworld} 
& 50.7 & 67.9 & 37.7 & 58.6 & \textbf{62.0} & 46.6 & 40.7 & 32.4 & 59.2 \\
\midrule
Ours & \textbf{63.2} & \textbf{68.9} & \textbf{47.3} & \textbf{64.9} & 61.2 & \textbf{62.1} & \textbf{82.4} & \textbf{44.9} & \textbf{73.8} \\
\bottomrule
\end{tabular*}

    }
\label{tab:vsibench}
\end{table}

\begin{table}[t]
    \centering
    \scriptsize
    \caption{
    \textbf{Evaluation of 3D QA} on SQA3D~\cite{ma2022sqa3d} and ScanQA~\cite{azuma2022scanqa}. \method{} achieves the best performance among 2D MLLMs on both benchmarks. Our method also outperforms most 3D MLLMs on SQA3D, demonstrating strong situated reasoning.}
    %
    \vspace{-0.2cm}
    \resizebox{\linewidth}{!}{\begin{tabular*}{\linewidth}{@{\extracolsep{\fill}} l c c c c c c @{}}
    \toprule
    \multirow{2}{*}{Model} &
    \multicolumn{2}{c}{SQA3D$_{\text{test}}$} &
    \multicolumn{4}{c}{ScanQA$_{\text{val}}$} \\
    \cmidrule(lr){2-3} \cmidrule(lr){4-7}
    & EM & EM-R & CIDEr & METEOR & ROUGE & EM \\
    \midrule
    
    \rowcolor{Light}
    \multicolumn{7}{l}{\textit{Expert Models}} \\
    SQA3D~\cite{ma2022sqa3d} & 46.6 & -- & -- & -- & -- & -- \\
    ScanQA~\cite{azuma2022scanqa} & -- & -- & 64.9 & 13.1 & 33.3 & 21.1 \\
    \midrule
    
    \rowcolor{Light}
    \multicolumn{7}{l}{\textit{3D MLLMs}} \\
    LEO~\cite{huang2024leo} & 50.0 & 52.4 & 80.0 & 16.2 & 39.3 & 21.5 \\
    Sig3D~\cite{man2024sig3d} & 52.6 & -- & 68.8 & 13.4 & 26.6 & -- \\
    View2Cap~\cite{yuan2025empoweringsituation} & 54.0 & 56.0 & 89.8 & 17.5 & 42.9 & 22.9 \\
    ChatScene~\cite{zhang2024chatscene} & 54.6 & 57.5 & 87.7 & 18.0 & 41.6 & 21.6 \\
    LLaVA-3D~\cite{zhu2024llava3d} & 55.6 & 57.6 & 91.7 & 20.7 & 50.1 & 27.0 \\
    3D-LLaVA~\cite{deng20253dllava} & 54.5 & 56.6 & 92.6 & 18.4 & 43.1 & -- \\
    Video-3D-LLM~\cite{zheng2024video3dllm} & 58.6 & -- & 102.1 & 20.0 & 49.3 & 30.1 \\
    3DRS~\cite{huang20253drs} & 60.6 & -- & 104.8 & 20.5 & 49.8 & 30.3 \\
    Ross3D~\cite{wang2025ross3d} & \textbf{63.0} & \textbf{65.7} & \textbf{107.0} & \textbf{20.9} & \textbf{50.7} & \textbf{30.8} \\
    \midrule
    
    \rowcolor{Light}
    \multicolumn{7}{l}{\textit{2D MLLMs}} \\
    SplatTalk~\cite{thai2025splattalk} & 47.6 & 49.4 & 77.5 & 15.6 & 38.5 & 22.4 \\
    SPAR~\cite{zhang2025spar} & 58.1 & -- & 90.7 & -- & -- & -- \\
    SpatialMLLM-4B~\cite{wu2025spatialmllmboostingmllmcapabilities} & 55.9 & 58.7 & 91.8 & 18.4 & 45.0 & 26.3 \\
    CdViews~\cite{wang20253dquestionanswering2d} & 56.9 & -- & 94.0 & -- & 46.8 & \textbf{30.1} \\
    Struct2D~\cite{zhu2025struct2d} & 58.5 & 61.3 & 92.1 & 17.4 & 44.1 & -- \\
    GPT4Scene~\cite{GPT4Scene} & 59.4 & 62.4 & 96.3 & 18.9 & 46.5 & 28.2 \\
    \midrule
    Ours & \textbf{62.8} & \textbf{65.0} & \textbf{100.4} & \textbf{19.5} & \textbf{47.9} & 28.2 \\
    \bottomrule
\end{tabular*}

    } 
    \label{table:scanqa}
    \vspace{-0.4cm}
\end{table}

\begin{table}[h]
    \centering
        \begin{minipage}[t]{0.48\textwidth}
        \caption{\textbf{Evaluation of 3D QA} on MSQA~\cite{linghu2024msr3d} (ScanNet). Methods that use the ground-truth location as model input are indicated with $^\dagger$. \method{} achieves the highest overall score.}
        \label{table:msr3d}
        \vspace{-0.3cm}
        \resizebox{\linewidth}{!}{%
            \begin{tabular}{l c c c c c c c}
    \toprule
    Model & \textit{Count.} & \textit{Exist.} & \textit{Attr.} & \textit{Spatial} & \textit{Navi.} & \textit{Others} & Overall \\
    \midrule
    SplatTalk~\cite{thai2025splattalk} & 19.6 & 60.3 & 44.0 & 35.8 & 35.5 & 61.8 & 41.8 \\
    GPT-4o~\cite{openai2024gpt4technicalreport} & 32.3 & 79.3 & \textbf{79.0} & 37.0 & 31.7 & \textbf{91.6} & 52.3 \\
    MSR3D$^\dagger$~\cite{linghu2024msr3d} & 32.3 & \textbf{93.1} & 50.0 & 46.5 & \textbf{54.1} & 75.6 & 54.2 \\
    LEO~\cite{huang2024leo} & 32.5 & 88.5 & 58.7 & 44.2 & 39.6 & 81.4 & 54.8 \\
    \midrule
    Ours & \textbf{33.1} & 88.5 & 61.3 & \textbf{57.6} & 47.2 & 83.8 & \textbf{58.6}\\
    \bottomrule
\end{tabular}%
        }
    \end{minipage}
    \hfill
    \begin{minipage}[t]{0.48\textwidth}
        \caption{\textbf{Evaluation of 3D QA} on Beacon3D~\cite{huang2025beacon3d} (ScanNet). \method{} achieves the highest overall score.}
        \label{table:beacon3d}
        \vspace{-0.16cm}
        \resizebox{\linewidth}{!}{%
            \begin{tabular}{lccccccc}
    \toprule
     Model & \textit{Class} & \textit{App.} & \textit{Geo.} & \textit{Spatial} & \textit{Exi.} & \makecell{Overall\\(Case)}
 & \makecell{Overall\\(Obj.)}
 \\
    \midrule
    SceneVerse~\cite{jia2024sceneverse} & 26.4 & 40.4 & 40.0 & 35.0 & 54.1 & 40.5 & 4.7 \\
    LEO~\cite{huang2024leo} & 16.4 & 39.8 & 47.6 & 52.8 & 54.3 & 45.2 & 7.5 \\
    Chat-Scene~\cite{zhang2024chatscene} & 36.4 & 39.8 & 56.7 & 47.6 & 48.8 & 45.8 & 7.8 \\
    GPT4Scene~\cite{GPT4Scene} & 38.1 & 59.7 & 59.3 & 52.6 & \textbf{66.1} & 57.2 & 17.9 \\
    LLaVA-3D~\cite{zhu2024llava3d} & 35.1 & 66.7 & \textbf{62.5} & 54.2 & 62.9 & 59.1 & 19.0 \\
    Video-3D LLM~\cite{zheng2024video3dllm} & 40.1 & 64.1 & 60.6 & 55.3 & 64.1 & 59.0 & 17.9 \\
    \midrule
    Ours & \textbf{44.8} & \textbf{66.8} & 55.4 & \textbf{64.7} & 65.4 & \textbf{62.4} & \textbf{23.4} \\
    \bottomrule
\end{tabular}
        }
    \end{minipage}
\end{table}

\paragraph{3D Question Answering Benchmarks.}
We evaluate \method{} on both situated 3D question answering and general 3D QA benchmarks. Our evaluation includes VSI-Bench which encompasses both query types. To assess situated reasoning, we use SQA3D~\cite{ma2022sqa3d} and MSQA~\cite{linghu2024msr3d}, which feature viewpoint-dependent question-answer pairs conditioned on natural language situation descriptions. For general 3D QA, we adopt ScanQA~\cite{azuma2022scanqa} and the zero-shot benchmark Beacon3D~\cite{huang2025beacon3d}. Following prior work~\cite{huang2024leo,thai2025splattalk}, we report MSQA and Beacon3D results on the ScanNet~\cite{dai2017scannet} split.

\paragraph{Evaluation Metrics.}
For VSI-Bench, we follow the official protocol by reporting Accuracy (ACC) for multiple-choice tasks and Mean Relative Accuracy (MRA) for numerical tasks. For ScanQA and SQA3D, we evaluate performance using standard n-gram-based language metrics. Specifically, ScanQA is assessed using CIDEr~\cite{cider2015}, METEOR~\cite{banerjee-lavie-2005-meteor}, ROUGE~\cite{lin-2004-rouge}, and EM~\cite{VQA}, while SQA3D uses EM and EM-Refined. For MSQA and Beacon3D, we adopt the evaluation protocols recommended by the respective authors and report GPT-based scores~\cite{OpenEQA2023}. Beacon3D further splits the overall metric into two categories: “Case” denoting overall accuracy, and “Obj” indicating object-centric accuracy, which requires correct predictions across all samples for a given object.

\paragraph{Comparison with other methods.}
Results on VSI-Bench are shown in Table~\ref{tab:vsibench}. VLM-3R~\cite{fan2025vlm3rvisionlanguagemodelsaugmented} is categorized as an expert model, as it is specifically optimized for VSI-Bench without training on other benchmarks and tasks. Among the remaining baselines, \method{} achieves the strongest overall performance by a significant margin. Our model excels particularly in tasks that require viewpoint understanding. This shows in substantial gains for Relative Direction ($+36.1\%$), Relative Distance ($+10.8\%$), and Route Planning ($+8.8\%$) compared to the second-best generalist baseline. These results demonstrate that our situation modeling module, supported by the global layout reconstruction objective, effectively enhance viewpoint-aware reasoning. By forming a persistent, cognitive map of the environment, our framework allows the model to accurately anchor the agent's situation and navigate complex egocentric spatial relations. For numerical tasks, \method{} obtains the best results in Absolute Distance and Object Size, highlighting the contribution of the camera-pose prior, which provides metric-scale cues crucial for accurate absolute reasoning.

Results on ScanQA and SQA3D are shown in~\cref{table:scanqa}. Following prior work~\cite{wang2025ross3d,thai2025splattalk,zheng2024video3dllm,wu2025spatialmllmboostingmllmcapabilities,wang20253dquestionanswering2d,zhu2024llava3d}, we group the methods into three categories. ``Expert models" are specifically fine-tuned for individual benchmarks without generalization to other tasks. ``3D MLLMs'' operate on point clouds or rely on explicit ground-truth depth and camera poses. ``2D MLLMs'' receive only image inputs, without any explicit 3D signals. Our method falls into the latter category. On SQA3D, our method achieves 62.8 EM, outperforming all 2D MLLMs and most 3D-based approaches. This highlights the effectiveness of our explicit situation modeling for robust viewpoint grounding and situated reasoning. For general QA on ScanQA, \method{} surpasses all other 2D methods across most metrics. Notably, across both benchmarks, our approach achieves the best performance among all methods capable of situation localization. 

Results on MSQA and Beacon3D are presented in~\cref{table:beacon3d} and~\cref{table:msr3d}, respectively. On both benchmarks, \method{} achieves the highest overall score with 58.6\% on MSQA and 62.4\% on Beacon3D. Notably, we observe pronounced gains in the \textit{spatial} subcategories of both benchmarks, outperforming the second-best methods by an absolute +11.1\% on MSQA and +9.4\% on Beacon3D.

\subsection{Ablation Studies}\label{sec:ablation}
We conduct ablation studies to analyze the impact of our proposed components in \method{}. All models in this section are trained on a smaller subset of the full training data. The metrics used are overall accuracy for VSI-Bench~\cite{yang2024thinkinginspace} and MSQA\cite{linghu2024msr3d}, CIDEr for ScanQA~\cite{azuma2022scanqa}, and EM for SQA3D~\cite{ma2022sqa3d}.

\begin{wraptable}{r}{0.48\textwidth}
    \centering
    \vspace{-1.2cm}
    \caption{\textbf{Ablations on Language-based Localization} on SQA3D~\cite{ma2022sqa3d}. Our full model achieves the best performance.}
    \resizebox{\linewidth}{!}{
        \begin{tabular}{rl cccc}
\toprule
& Method & Acc@0.5m & Acc@1.0m & Acc@15\textdegree & Acc@30\textdegree \\
\midrule
{\scriptsize{1}} & Situation & 27.0 & 51.5 & 26.7 & 48.7 \\
{\scriptsize{2}} & \textcircled{\scriptsize{1}} + Layout & 30.1 & 59.3 & 28.2 & 53.2 \\
{\scriptsize{3}} & \textcircled{\scriptsize{1}} + \textcircled{\scriptsize{2}} + Cam & \textbf{39.9} & \textbf{75.5} & \textbf{31.9} & \textbf{56.3} \\
\bottomrule
\end{tabular}

    }
    \vspace{-0.775cm}
    \label{tabl:abl_loc}
\end{wraptable}

\paragraph{Effectiveness of Components on Localization}
As shown in~\cref{tabl:abl_loc}, explicitly modeling the situation via our query tokens already provides a strong baseline for localization, enabling the model to infer position and orientation directly from language (Row 1).
Adding the layout reconstruction objective further improves localization accuracy  (Row 2). This confirms that the BEV objective helps the model consolidate multi-view video observations into a coherent global map, supplying more stable anchors and strengthening the fine-grained relational cues needed for accurate spatial grounding.
Finally, integrating the camera pose prior leads to another large improvement (Row 3). The pose prior complements the spatial supervision modules by injecting pose cues into the token stream: while layout reconstruction contributes global consistency and situation modeling supports egocentric awareness, the camera prior ensures that the internal map is accurately aligned in metric space. This additional pose information from the 3D foundation model particularly benefits position estimation, which requires predicting metric-scale absolute coordinates of the agent.

\begin{wraptable}{r}{0.48\textwidth}
    \centering
    \vspace{-1.2cm}
    \caption{\textbf{Ablations on 3D QA}. Consistent gains from each component validate their individual utility, while the full model's peak performance confirms the complementary nature and effectiveness of our framework.}
    \resizebox{\linewidth}{!}{%
        \begin{tabular}{rl cccc}
    \toprule
        & Method & VSI-Bench & ScanQA & SQA3D & MSQA \\
    \midrule
        {\scriptsize{1}} & LLaVA FT 
            & 49.9 & 92.2 & 58.4 & 54.4 \\
        {\scriptsize{2}} & \textcircled{\scriptsize{1}} + Situation 
            & 50.6 & 98.4 & 59.2 & 55.1 \\
        {\scriptsize{3}} & \textcircled{\scriptsize{1}} + Layout 
            & 50.3 & 99.7 & 58.9 & 55.5 \\
        {\scriptsize{4}} & \textcircled{\scriptsize{1}} + \textcircled{\scriptsize{2}} + \textcircled{\scriptsize{3}} 
            & 53.6 & 104.3 & 59.6 & 56.0 \\
        {\scriptsize{5}} & \textcircled{\scriptsize{4}} + Cam 
            & \textbf{54.3} & \textbf{107.3} & \textbf{59.9} & \textbf{56.6} \\
    \bottomrule
\end{tabular}
    }
    \label{tbl:abl_qa}
    \vspace{-0.775cm}
\end{wraptable}

\paragraph{Effectiveness of Components on 3D QA.}
As shown in \cref{tbl:abl_qa}, introducing situation modeling (Row 2) yields consistent gains on SQA3D and MSQA over the fine-tuned base model LLaVA-Video-7B~\cite{zhang2024llava} (Row 1). This indicates that explicitly encoding the agent’s pose helps the model resolve viewpoint-dependent references and spatial relations. Notably, we also observe clear improvements on ScanQA, which does not provide situation descriptions. This suggests that the situation module not only supports situational reasoning but also strengthens the model’s general spatial understanding: by learning to predict position and orientation from text, the model forms relational awareness even when no explicit localization query is present. Adding global layout reconstruction on top of the base model similarly improves performance (Row 3). When combined (Row 4), the two components yield larger gains than either alone, underscoring their complementary roles: situation modeling grounds a local viewpoint, while layout reconstruction organizes the global scene. Finally, incorporating camera pose priors to obtain our final model achieves the strongest performance (Row 5). 
While the external camera tokens yield a larger gain in the localization task, their additional improvement in QA is comparatively modest. This suggests that 3D QA relies predominantly on the relational and global scene understanding established by our spatial training objectives, whereas localization benefits more directly from the strong geometric features for metric-accurate coordinate regression.

\begin{wraptable}{r}{0.48\textwidth}
    \centering
    \vspace{-1.2cm}
    \caption{\textbf{Ablations on 3D Foundation Model Feature Selection.} Integrating only the CUT3R camera token as a spatial prior proves superior to including both camera and geometry tokens.}
    \resizebox{\linewidth}{!}{%
        \begin{tabular}{lcccc}
\toprule
\multirow{2}{*}{3D Features} &
VSI-Bench &
\multicolumn{3}{c}{SQA3D} \\
\cmidrule(lr){2-2} \cmidrule(lr){3-5}
 & Avg. & EM & Acc@1m & Acc@30$^\circ$ \\
\midrule
Cam.\ \texttt{+} Geom. 
    & 59.5 & 59.0 & 71.8 & 59.8 \\
Cam.\ (Ours) 
    & \textbf{63.2} & \textbf{62.8} & \textbf{75.9} & \textbf{63.0} \\
\bottomrule
\end{tabular}%
    }
    \label{tbl:abl_cut3r}
    \vspace{-0.5cm}
\end{wraptable}

\paragraph{3D Foundation Model Feature Selection.} 
As outlined in~\cref{sec:cam_token}, CUT3R~\cite{wang2025cut3r} produces two types of latent tokens for every frame: a camera token $z'_t$ and geometry tokens $F'_t$. In \method{}, we use only the camera token $z'_t$ as the geometric prior. Our design is guided by two considerations:
(1) the camera token alone provides a sufficiently strong pose cue for grounding video frames in 3D, and (2) avoiding early fusion of geometry tokens with the visual embeddings preserves the integrity of the pre-trained vision–language feature space. To evaluate our choice, we train a variant of our model which incorporates both token types. In addition to our camera token integration strategy described in~\cref{{sec:cam_token}}, we follow a similar approach to~\cite{wu2025spatialmllmboostingmllmcapabilities} and fuse geometry and vision tokens via MLP projection and token-wise addition. As shown in~\cref{tbl:abl_cut3r}, using only the camera token outperforms the combined approach. This confirms our hypothesis: the camera token is sufficient as a spatial prior, whereas additional geometry tokens likely introduce redundant signals that interfere with pre-trained representations.

\section{Conclusion}\label{sec:conclusion}
We have presented \method{}, a framework that equips a 2D Vision-Language Model (VLM) with advanced 3D understanding capabilities. Operating on video data without explicit 3D input, \method{} is able to perform language-based localization and spatial reasoning tasks. Inspired by principles of human cognition, we develop two complementary modules that jointly enhance global scene understanding and situational awareness. In addition, we incorporate lightweight pose priors from a 3D foundation model, enabling metric-scale consistency. Our experiments show that combining explicit spatial supervision with geometric priors yields substantial gains across language-driven localization, situated reasoning, and general 3D question-answering tasks. These results highlight that robust 3D understanding can emerge directly from video input when models are guided to organize visual information into global and viewpoint-aware representations. We hope that \method{} inspires further research toward 3D-aware VLMs with stronger spatial and embodied understanding.


\bibliographystyle{splncs04}
\bibliography{main}

@String(CVPR  = {IEEE Conf. Comput. Vis. Pattern Recog.})

@String(ICCV  = {Int. Conf. Comput. Vis.})

@String(ECCV  = {Eur. Conf. Comput. Vis.})

@String(NeurIPS = {Adv. Neural Inform. Process. Syst.})

@String(ICML  = {Int. Conf. Mach. Learn.})

@String(ICLR  = {Int. Conf. Learn. Represent.})

@String(AAAI  = {AAAI})

@String(CVPR  = {CVPR})

@String(ICCV  = {ICCV})

@String(ECCV  = {ECCV})

@String(NeurIPS = {NeurIPS})

@String(ICML  = {ICML})

@String(ICLR  = {ICLR})

@article{zhang2024llava,
    title={Video Instruction Tuning With Synthetic Data}, 
    author={Yuanhan Zhang and Jinming Wu and Wei Li and Bo Li and Zejun Ma and Ziwei Liu and Chunyuan Li},
    year={2024},
    journal = {arXiv preprint arXiv:2410.02713}, 
}

@inproceedings{azuma2022scanqa,
  title={ScanQA: 3D Question Answering for Spatial Scene Understanding},
  author={Azuma, Daichi and Miyanishi, Taiki and Kurita, Shuhei and Kawanabe, Motoaki},
  booktitle={CVPR},
  year={2022}
}

@inproceedings{ma2022sqa3d,
  title={SQA3D: Situated Question Answering in 3D Scenes},
  author={Ma, Xiaojian and Yong, Silong and Zheng, Zilong and Li, Qing and Liang, Yitao and Zhu, Song-Chun and Huang, Siyuan},
  booktitle={ICLR},
  year={2023}
}

@inproceedings{linghu2024msr3d,
  author    = {Linghu, Xiongkun and Huang, Jiangyong and Niu, Xuesong and Ma, Xiaojian and Jia, Baoxiong and Huang, Siyuan},
  booktitle = {NeurIPS},
  title     = {Multi-modal Situated Reasoning in 3D Scenes},
  year      = {2024}
}

@inproceedings{huang2025beacon3d,
  title={Unveiling the Mist over 3D Vision-Language Understanding: Object-centric Evaluation with Chain-of-Analysis},
  author={Huang, Jiangyong and Jia, Baoxiong and Wang, Yan and Zhu, Ziyu and Linghu, Xiongkun and Li, Qing and Zhu, Song-Chun and Huang, Siyuan},
  booktitle={CVPR},
  year={2025}
}

@inproceedings{zhang2025spar,
            title={From Flatland to Space: Teaching Vision-Language Models to Perceive and Reason in 3D},
            author={Zhang, Jiahui and Chen, Yurui and Zhou, Yanpeng and Xu, Yueming and Huang, Ze and Mei, Jilin and Chen, Junhui and Yuan, Yujie and Cai, Xinyue and Huang, Guowei and Quan, Xingyue and Xu, Hang and Zhang, Li},
            year={2025},
            booktitle={NeurIPS},
          }

@inproceedings{hong20233dllm,
 author = {Hong, Yining and Zhen, Haoyu and Chen, Peihao and Zheng, Shuhong and Du, Yilun and Chen, Zhenfang and Gan, Chuang},
 title = {3D-LLM: Injecting the 3D World into Large Language Models},
 booktitle = {NeurIPS},
 year = {2023},
}

@inproceedings{chen2023ll3da,
    title={LL3DA: Visual Interactive Instruction Tuning for Omni-3D Understanding, Reasoning, and Planning}, 
    author={Sijin Chen and Xin Chen and Chi Zhang and Mingsheng Li and Gang Yu and Hao Fei and Hongyuan Zhu and Jiayuan Fan and Tao Chen},
    year={2024},
    booktitle = {CVPR},
}

@inproceedings{huang2024leo,
  title={An Embodied Generalist Agent in 3D World},
  author={Huang, Jiangyong and Yong, Silong and Ma, Xiaojian and Linghu, Xiongkun and Li, Puhao and Wang, Yan and Li, Qing and Zhu, Song-Chun and Jia, Baoxiong and Huang, Siyuan},
  booktitle={ICML},
  year={2024}
}

@inproceedings{zhang2024chatscene,
  title={ChatScene: Knowledge-Enabled Safety-Critical Scenario Generation for Autonomous Vehicles},
  author={Zhang, Jiawei and Xu, Chejian and Li, Bo},
  booktitle={CVPR},
  year={2024}
}

@inproceedings{zhu2024llava3d,
  title={LLaVA-3D: A Simple yet Effective Pathway to Empowering LMMs with 3D-awareness},
  author={Zhu, Chenming and Wang, Tai and Zhang, Wenwei and Pang, Jiangmiao and Liu, Xihui},
  booktitle={ICCV},
  year={2025}
}

@inproceedings{zheng2024video3dllm,
      title={Video-3D LLM: Learning Position-Aware Video Representation for 3D Scene Understanding}, 
      author={Duo Zheng and Shijia Huang and Liwei Wang},
      year={2025},
      booktitle={CVPR},
}

@inproceedings{huang20253drs,
  title={MLLMs Need 3D-Aware Representation Supervision for Scene Understanding},
  author={Xiaohu Huang and Jingjing Wu and Qunyi Xie and Kai Han},
  booktitle={NeurIPS},
  year={2025}
}

@inproceedings{wang2025ross3d,
  title={Ross3D: Reconstructive visual instruction tuning with 3D-awareness},
  author={Wang, Haochen and Zhao, Yucheng and Wang, Tiancai and Fan, Haoqiang and Zhang, Xiangyu and Zhang, Zhaoxiang},
  booktitle={ICCV},
  year={2025}
}

@article{chen2024internvl2,
  title={Expanding Performance Boundaries of Open-Source Multimodal Models with Model, Data, and Test-Time Scaling},
  author={Chen, Zhe and Wang, Weiyun and Cao, Yue and Liu, Yangzhou and Gao, Zhangwei and Cui, Erfei and Zhu, Jinguo and Ye, Shenglong and Tian, Hao and Liu, Zhaoyang and others},
  journal={arXiv preprint arXiv:2412.05271},
  year={2024}
}

@article{Qwen2-VL,
  title={Qwen2-VL: Enhancing Vision-Language Model's Perception of the World at Any Resolution},
  author={Wang, Peng and Bai, Shuai and Tan, Sinan and Wang, Shijie and Fan, Zhihao and Bai, Jinze and Chen, Keqin and Liu, Xuejing and Wang, Jialin and Ge, Wenbin and Fan, Yang and Dang, Kai and Du, Mengfei and Ren, Xuancheng and Men, Rui and Liu, Dayiheng and Zhou, Chang and Zhou, Jingren and Lin, Junyang},
  journal={arXiv preprint arXiv:2409.12191},
  year={2024}
}

@inproceedings{thai2025splattalk,
  title={Splattalk: 3d vqa with gaussian splatting},
  author={Thai, Anh and Peng, Songyou and Genova, Kyle and Guibas, Leonidas and Funkhouser, Thomas},
  booktitle={ICCV},
  year={2025}
}

@inproceedings{wu2025spatialmllmboostingmllmcapabilities,
  title={Spatial-MLLM: Boosting MLLM Capabilities in Visual-based Spatial Intelligence},
  author={Wu, Diankun  and Liu, Fangfu and Hung, Yi-Hsin and Duan, Yueqi},
  booktitle={NeurIPS},
  year={2025}
}

@article{GPT4Scene,
  title={GPT4Scene: Understand 3D Scenes from Videos with Vision-Language Models},
  author={Zhangyang Qi and Zhixiong Zhang and Ye Fang and Jiaqi Wang and Hengshuang Zhao},
  journal={arXiv preprint arXiv:2501.01428},
  year={2024}
}

@inproceedings{zhu20233dvista,
  title={3D-VisTA: Pre-trained Transformer for 3D Vision and Text Alignment},
  author={Zhu, Ziyu and Ma, Xiaojian and Chen, Yixin and Deng, Zhidong and Huang, Siyuan and Li, Qing},
  booktitle={ICCV},
  year={2023}
}

@inproceedings{zhu20pq3d,
      title={Unifying 3D Vision-Language Understanding via Promptable Queries},
      author={Zhu, Ziyu and Zhang, Zhuofan and Ma, Xiaojian and Niu, Xuesong and Chen, Yixin and Jia, Baoxiong and Deng, Zhidong and Huang, Siyuan and Li, Qing},
      booktitle={ECCV},
      year={2024}
}

@inproceedings{jia2024sceneverse,
      title={Sceneverse: Scaling 3d vision-language learning for grounded scene understanding},
      author={Jia, Baoxiong and Chen, Yixin and Yu, Huangyue and Wang, Yan and Niu, Xuesong and Liu, Tengyu and Li, Qing and Huang, Siyuan},
      booktitle={ECCV},
      year={2024}
    }

@article{openai2024gpt4technicalreport,
      title={GPT-4 Technical Report}, 
      author={OpenAI},
      year={2024},
      journal={arXiv preprint arXiv:2303.08774}, 
}

@inproceedings{man2024sig3d,
      title={Situational Awareness Matters in 3D Vision Language Reasoning},
      author={Man, Yunze and Gui, Liang-Yan and Wang, Yu-Xiong},
      booktitle={CVPR},
      year={2024}
}

@inproceedings{yuan2025empoweringsituation,
      title={Empowering Large Language Models with 3D Situation Awareness}, 
      author={Zhihao Yuan and Yibo Peng and Jinke Ren and Yinghong Liao and Yatong Han and Chun-Mei Feng and Hengshuang Zhao and Guanbin Li and Shuguang Cui and Zhen Li},
      year={2025},
      booktitle={CVPR}, 
}

@inproceedings{wang20253dquestionanswering2d,
      title={3D Question Answering via only 2D Vision-Language Models},
      author={Fengyun Wang and Sicheng Yu and Jiawei Wu and Jinhui Tang and Hanwang Zhang and Qianru Sun},
      booktitle={ICML},
      year={2025}
}

@inproceedings{deng20253dllava,
  title={3D-LLaVA: Towards Generalist 3D LMMs with Omni Superpoint Transformer},
  author={Deng, Jiajun and He, Tianyu and Jiang, Li and Wang, Tianyu and Dayoub, Feras and Reid, Ian},
  booktitle={CVPR},
  year={2025}
}

@article{comanici2025gemini25pushingfrontier,
      title={Gemini 2.5: Pushing the Frontier with Advanced Reasoning, Multimodality, Long Context, and Next Generation Agentic Capabilities}, 
      author={Gemini Team},
      year={2025},
      journal={arXiv preprint arXiv:2507.06261}, 
}

@inproceedings{zhu2025struct2d,
  title={Struct2D: A Perception-Guided Framework for Spatial Reasoning in MLLMs},
  author={Zhu, Fangrui and Wang, Hanhui and Xie, Yiming and Gu, Jing and Ding, Tianye and Yang, Jianwei and Jiang, Huaizu},
  booktitle={NeurIPS},
  year={2025}
}

@inproceedings{wang2025cut3r,
  title={Continuous 3D Perception Model with Persistent State},
  author={Wang, Qianqian and Zhang, Yifei and Holynski, Aleksander and Efros, Alexei A and Kanazawa, Angjoo},
  booktitle={CVPR},
  year={2025}
}

@article{zhang2025mllmsstrugglespatialunderstanding,
      title={Why Do MLLMs Struggle with Spatial Understanding? A Systematic Analysis from Data to Architecture}, 
      author={Wanyue Zhang and Yibin Huang and Yangbin Xu and JingJing Huang and Helu Zhi and Shuo Ren and Wang Xu and Jiajun Zhang},
      year={2025},
      journal={arXiv preprint arXiv:2509.02359}, 
}

@inproceedings{yang2024thinkinginspace,
    title={{Thinking in Space: How Multimodal Large Language Models See, Remember and Recall Spaces}},
    author={Yang, Jihan and Yang, Shusheng and Gupta, Anjali W. and Han, Rilyn and Fei-Fei, Li and Xie, Saining},
    booktitle={CVPR},
    year={2025}
}

@inproceedings{kamath2023whatsup,
  title={What's ``up'' with vision-language models? Investigating their struggle with spatial reasoning},
  author={Kamath, Amita and Hessel, Jack and Chang, Kai-Wei},
  booktitle={EMNLP},
  year={2023}
}

@inproceedings{chen2025spatialreasoninghardvlms,
      title={Why Is Spatial Reasoning Hard for VLMs? An Attention Mechanism Perspective on Focus Areas}, 
      author={Shiqi Chen and Tongyao Zhu and Ruochen Zhou and Jinghan Zhang and Siyang Gao and Juan Carlos Niebles and Mor Geva and Junxian He and Jiajun Wu and Manling Li},
      booktitle={ICML},
      year={2025}
}

@article{fan2025vlm3rvisionlanguagemodelsaugmented,
      title={VLM-3R: Vision-Language Models Augmented with Instruction-Aligned 3D Reconstruction}, 
      author={Zhiwen Fan and Jian Zhang and Renjie Li and Junge Zhang and Runjin Chen and Hezhen Hu and Kevin Wang and Huaizhi Qu and Dilin Wang and Zhicheng Yan and Hongyu Xu and Justin Theiss and Tianlong Chen and Jiachen Li and Zhengzhong Tu and Zhangyang Wang and Rakesh Ranjan},
      year={2025},
      journal={arXiv preprint arXiv:2505.20279}
}

@article{zheng2025learningvideos3dworld,
      title={Learning from Videos for 3D World: Enhancing MLLMs with 3D Vision Geometry Priors}, 
      author={Duo Zheng and Shijia Huang and Yanyang Li and Liwei Wang},
      year={2025},
      journal={arXiv preprint arXiv:2505.24625}, 
}

@inproceedings{wang2025vggt,
  title={VGGT: Visual Geometry Grounded Transformer},
  author={Wang, Jianyuan and Chen, Minghao and Karaev, Nikita and Vedaldi, Andrea and Rupprecht, Christian and Novotny, David},
  booktitle={CVPR},
  year={2025}
}

@inproceedings{dai2017scannet,
    title={ScanNet: Richly-annotated 3D Reconstructions of Indoor Scenes},
    author={Dai, Angela and Chang, Angel X. and Savva, Manolis and Halber, Maciej and Funkhouser, Thomas and Nie{\ss}ner, Matthias},
    booktitle = {CVPR},
    year = {2017}
}

@article{huang2023chat3dv2,
  title={Chat-3D v2: Bridging 3D Scene and Large Language Models with Object Identifiers},
  author={Huang, Haifeng and Wang, Zehan and Huang, Rongjie and Liu, Luping and Cheng, Xize and Zhao, Yang and Jin, Tao and Zhao, Zhou},
  journal={arXiv preprint arXiv:2312.08168},
  year={2023}
}

@inproceedings{fu2024scenellmextendinglanguagemodel,
      title={Scene-LLM: Extending Language Model for 3D Visual Understanding and Reasoning}, 
      author={Rao Fu and Jingyu Liu and Xilun Chen and Yixin Nie and Wenhan Xiong},
      booktitle = {WACV},
      year={2025},
}

@inproceedings{driess2023palme,
    title={PaLM-E: An Embodied Multimodal Language Model},
    author={Driess, Danny and Xia, Fei and Sajjadi, Mehdi S. M. and Lynch, Corey and Chowdhery, Aakanksha and Ichter, Brian and Wahid, Ayzaan and Tompson, Jonathan and Vuong, Quan and Yu, Tianhe and Huang, Wenlong and Chebotar, Yevgen and Sermanet, Pierre and Duckworth, Daniel and Levine, Sergey and Vanhoucke, Vincent and Hausman, Karol and Toussaint, Marc and Greff, Klaus and Zeng, Andy and Mordatch, Igor and Florence, Pete},
    booktitle={ICML},
    year={2023}
}

@article{open_x_embodiment_rt_x_2023,
title={Open {X-E}mbodiment: Robotic Learning Datasets and {RT-X} Models},
author = {Open X-Embodiment Collaboration },
journal = {arXiv preprint arXiv:2310.08864},
year = {2023},
}

@inbook{Newcombe2024Spatial,
	author = {Newcombe, Nora S.},
	booktitle = {Open {Encyclopedia} of {Cognitive} {Science}},
	editor = {Frank, Michael C. and Majid, Asifa},
	year = {2024},
	month = {jul 24},
	note = {https://oecs.mit.edu/pub/or750iar},
	publisher = {MIT Press},
	title = {Spatial {Cognition}},
}

@article{kendall2017uncertainties,
  title={What uncertainties do we need in bayesian deep learning for computer vision?},
  author={Kendall, Alex and Gal, Yarin},
  journal=NIPS,
  year={2017}
}

@article{Tolman1948,
  author       = {Tolman, E. C.},
  title        = {Cognitive Maps in Rats and Men},
  journal      = {Psychological Review},
  year         = {1948},
  volume       = {55},
  number       = {4},
  pages        = {189--208},
  doi          = {10.1037/h0061626}
}

@inproceedings{yu2025inst3dllm,
  title={Inst3d-lmm: Instance-aware 3d scene understanding with multi-modal instruction tuning},
  author={Yu, Hanxun and Li, Wentong and Wang, Song and Chen, Junbo and Zhu, Jianke},
  booktitle={CVPR},
  year={2025}
}

@inproceedings{zhi2024lscenellm,
  title={LSceneLLM: Enhancing Large 3D Scene Understanding Using Adaptive Visual Preferences},
  author={Zhi, Hongyan and Chen, Peihao and Li, Junyan and Ma, Shuailei and Sun, Xinyu and Xiang, Tianhang and Lei, Yinjie and Tan, Mingkui and Gan, Chuang},
  booktitle={CVPR},
  year={2025}
}

@inproceedings{procthor,
  author={Matt Deitke and Eli VanderBilt and Alvaro Herrasti and
          Luca Weihs and Jordi Salvador and Kiana Ehsani and
          Winson Han and Eric Kolve and Ali Farhadi and
          Aniruddha Kembhavi and Roozbeh Mottaghi},
  title={{ProcTHOR: Large-Scale Embodied AI Using Procedural Generation}},
  booktitle={NeurIPS},
  year={2022},
}

@inproceedings{rt22023arxiv,
    title={RT-2: Vision-Language-Action Models Transfer Web Knowledge to Robotic Control},
    author={Anthony Brohan and Noah Brown and Justice Carbajal and Yevgen Chebotar and Xi Chen and Krzysztof Choromanski and Tianli Ding and Danny Driess and Avinava Dubey and Chelsea Finn and et al.},
    booktitle={CoRL},
    year={2023}
}

@inproceedings{tian2024DriveVLM,
    title={DriveVLM: The Convergence of Autonomous Driving and Large Vision-Language Models},
    author={Xiaoyu Tian and Junru Gu and Bailin Li and Yicheng Liu and Zhiyong Zhao and Yang Wang and Kun Zhan and Peng Jia and Xianpeng Lang and Hang Zhao},
    booktitle={CoRL},
    year={2024}
}

@inproceedings{Kong_2025_vlrdriver,
    author    = {Kong, Fanjie and Li, Yitong and Chen, Weihuang and Min, Chen and Li, Yizhe and Gao, Zhiqiang and Li, Haoyang and Guo, Zhongyu and Sun, Hongbin},
    title     = {VLR-Driver: Large Vision-Language-Reasoning Models for Embodied Autonomous Driving},
    year = {2025},
    booktitle = {ICCV}
}

@inproceedings{ma2024dolphins,
author = {Ma, Yingzi and Cao, Yulong and Sun, Jiachen and Pavone, Marco and Xiao, Chaowei},
title = {Dolphins: Multimodal Language Model for Driving},
year = {2024},
booktitle = ECCV,
}

@inproceedings{cider2015,
  author={Vedantam, Ramakrishna and Zitnick, C. Lawrence and Parikh, Devi},
  booktitle={CVPR}, 
  title={CIDEr: Consensus-based image description evaluation}, 
  year={2015}
}

@inproceedings{lin-2004-rouge,
    title = "{ROUGE}: A Package for Automatic Evaluation of Summaries",
    author = "Lin, Chin-Yew",
    booktitle = "Text Summarization Branches Out",
    year = "2004"
}

@inproceedings{banerjee-lavie-2005-meteor,
    title ={{METEOR}: An Automatic Metric for {MT} Evaluation with Improved Correlation with Human Judgments},
    author = {Banerjee, Satanjeev  and Lavie, Alon},
    booktitle = {ACL Workshop on Intrinsic and Extrinsic Evaluation Measures for Machine Translation and/or Summarization},
    year = {2005}
}

@inproceedings{VQA,
    author = {Stanislaw Antol and Aishwarya Agrawal and Jiasen Lu and Margaret Mitchell and Dhruv Batra and C. Lawrence Zitnick and Devi Parikh},
    title = {{VQA}: {V}isual {Q}uestion {A}nswering},
    booktitle = {ICCV},
    year = {2015}
}

@inproceedings{OpenEQA2023,
    title         = {OpenEQA: Embodied Question Answering in the Era of Foundation Models}, 
    booktitle     = {CVPR},
    author        = {Majumdar, Arjun and Ajay, Anurag and Zhang, Xiaohan and Putta, Pranav and Yenamandra, Sriram and Henaff, Mikael and Silwal, Sneha and et al.},
    year          = {2024},
}

@inproceedings{wang2023ret,
      title={Text to Point Cloud Localization with Relation-Enhanced Transformer}, 
      author={Guangzhi Wang and Hehe Fan and Mohan Kankanhalli},
      year={2023},
      booktitle={AAAI},
}

@inproceedings{xia2024text2loc,
      title={Text2Loc: 3D Point Cloud Localization from Natural Language},
      author={Xia, Yan and Shi, Letian and Ding, Zifeng and Henriques, Jo{\~a}o F and Cremers, Daniel},
      booktitle={CVPR},
      year={2024}
}

@article{wang2024instancefreetextpointcloud,
      title={Instance-free Text to Point Cloud Localization with Relative Position Awareness}, 
      author={Lichao Wang and Zhihao Yuan and Jinke Ren and Shuguang Cui and Zhen Li},
      year={2024},
      journal={arXiv preprint arXiv:2404.17845}, 
}

@inproceedings{xu2025cmmlocadvancingtexttopointcloudlocalization,
      title={CMMLoc: Advancing Text-to-PointCloud Localization with Cauchy-Mixture-Model Based Framework}, 
      author={Yanlong Xu and Haoxuan Qu and Jun Liu and Wenxiao Zhang and Xun Yang},
      year={2025},
      booktitle={CVPR},
}

@inproceedings{huang2024chatscene,
  title={Chat-scene: Bridging 3d scene and large language models with object identifiers},
  author={Huang, Haifeng and Chen, Yilun and Wang, Zehan and Huang, Rongjie and Xu, Runsen and Wang, Tai and Liu, Luping and Cheng, Xize and Zhao, Yang and Pang, Jiangmiao and others},
  year={2024},
  booktitle={NeurIPS},
}

@inproceedings{kang2024robin3d,
  title={Robin3d: Improving 3d large language model via robust instruction tuning},
  author={Kang, Weitai and Huang, Haifeng and Shang, Yuzhang and Shah, Mubarak and Yan, Yan},
  booktitle={ICCV},
  year={2025}
}

@article{Bottini2020KnowledgeAR,
  title={Knowledge Across Reference Frames: Cognitive Maps and Image Spaces},
  author={Roberto Bottini and Christian F. Doeller},
  journal={Trends in Cognitive Sciences},
  year={2020},
  volume={24},
  pages={606-619},
}

@inproceedings{lee2025perspective,
title  = {Perspective-Aware Reasoning in Vision-Language Models via Mental Imagery Simulation},
author = {Lee, Phillip Y. and Je, Jihyeon and Park, Chanho and Uy, Mikaela Angelina and Guibas, Leonidas and Sung, Minhyuk},
booktitle = {ICCV},
year   = {2025}
}

@inproceedings{
  zhang2025do,
  title={Do Vision-Language Models Represent Space and How? Evaluating Spatial Frame of Reference under Ambiguities},
  author={Zheyuan Zhang and Fengyuan Hu and Jayjun Lee and Freda Shi and Parisa Kordjamshidi and Joyce Chai and Ziqiao Ma},
  booktitle={ICML},
  year={2025},
}

@article{goral2024seeingeyesevaluatingvisual,
  title={Seeing Through Their Eyes: Evaluating Visual Perspective Taking in Vision Language Models}, 
  author={Gracjan Góral and Alicja Ziarko and Michal Nauman and Maciej Wołczyk},
  year={2024},
  journal={arXiv preprint arXiv:2409.12969}
}

@article{kosslyn1978,
author = {Kosslyn, Stephen and Ball, Thomas and Reiser, Brian},
year = {1978},
title = {Visual images preserve metric spatial information: Evidence from studies of image scanning},
journal = {Journal of Experimental Psychology: Human Perception and Performance},
}

@inproceedings{sep-mental-imagery,
	author       =	{Nanay, Bence},
	title        =	{{Mental Imagery}},
	booktitle    =	{The {Stanford} Encyclopedia of Philosophy},
	year         =	{2021},
}

@book{Paivio1979,
  author    = {Paivio, Allan},
  title     = {Imagery and Verbal Processes},
  edition   = {1},
  year      = {1979},
  publisher = {Psychology Press},
}

@article{zhang2025sphere,
    author = {Zhang, Wenyu and Ng, Wei En and Ma, Lixin and Wang, Yuwen and Zhao, Junqi and Koenecke, Allison and Li, Boyang and Wang, Lu},
    year = {2025},
    title = {{SPHERE: Unveiling Spatial Blind Spots in Vision-Language Models Through Hierarchical Evaluation}},
    journal = {ACL},
}

@article{cheng20253dawareregionprompted,
      title={3D Aware Region Prompted Vision Language Model}, 
      author={An-Chieh Cheng and Yang Fu and Yukang Chen and Zhijian Liu and Xiaolong Li and Subhashree Radhakrishnan and Song Han and Yao Lu and Jan Kautz and Pavlo Molchanov and Hongxu Yin and Xiaolong Wang and Sifei Liu},
      year={2025},
      journal={arXiv preprint arXiv:2509.13317}, 
}

@article{xu2025multi,
  title={Multi-SpatialMLLM: Multi-Frame Spatial Understanding with Multi-Modal Large Language Models},
  author={Xu, Runsen and Wang, Weiyao and Tang, Hao and Chen, Xingyu and Wang, Xiaodong and Chu, Fu-Jen and Lin, Dahua and Feiszli, Matt and Liang, Kevin J.},
  journal={arXiv preprint arXiv:2505.17015},
  year={2025}
}

@inproceedings{liu2024mmbench,
  title={Mmbench: Is your multi-modal model an all-around player?},
  author={Liu, Yuan and Duan, Haodong and Zhang, Yuanhan and Li, Bo and Zhang, Songyang and Zhao, Wangbo and Yuan, Yike and Wang, Jiaqi and He, Conghui and Liu, Ziwei and others},
  booktitle={ECCV},
  year={2024}
}

@inproceedings{wu2025reinforcingspatialreasoningvisionlanguage,
  title={Reinforcing Spatial Reasoning in Vision-Language Models with Interwoven Thinking and Visual Drawing}, 
  author={Junfei Wu and Jian Guan and Kaituo Feng and Qiang Liu and Shu Wu and Liang Wang and Wei Wu and Tieniu Tan},
  booktitle={NeurIPS},
  year={2025},
}

@article{ray2025satdynamicspatialaptitude,
      title={SAT: Dynamic Spatial Aptitude Training for Multimodal Language Models}, 
      author={Arijit Ray and Jiafei Duan and Ellis Brown and Reuben Tan and Dina Bashkirova and Rose Hendrix and Kiana Ehsani and Aniruddha Kembhavi and Bryan A. Plummer and Ranjay Krishna and Kuo-Hao Zeng and Kate Saenko},
      year={2025},
      journal={arXiv preprint arXiv:2412.07755}, 
    }

@inproceedings{zhai2023sigmoid,
  title={Sigmoid loss for language image pre-training},
  author={Zhai, Xiaohua and Mustafa, Basil and Kolesnikov, Alexander and Beyer, Lucas},
  booktitle={ICCV},
  year={2023}
}
 
\newpage
\appendix
\setcounter{section}{0}
\counterwithin{table}{section}
\counterwithin{figure}{section}

In the supplementary material, we provide:
\begin{itemize}
    \item Training data statistics and evaluation details (\cref{sec:add_impl_details})
    \item Analysis of robustness to the choice of 3D foundation model (\cref{sec:vggt})
    \item Inference efficiency analysis (\cref{sec:efficiency})
    \item Analysis of the correlation of localization accuracy with situated QA performance and uncertainty (\cref{sec:effects_loc})
    \item Details of the Bird's-Eye-View (BEV) representation (\cref{sec:bev})
    \item Evaluation and ablation of global layout reconstruction (\cref{sec:layout_acc})
    \item Frame count analysis (\cref{sec:frame_sampling})
    \item Additional qualitative results (\cref{sec:supp_qualitative})
    \item Failure case visualizations (\cref{sec:supp_failures})
    \item Discussion of limitations and future work (\cref{sec:supp_limitations})
\end{itemize}

\section{Training Data and Evaluation Details}\label{sec:add_impl_details}
\paragraph{Training Data Statistics.}
We report detailed training data statistics in~\cref{tab:dataset}. For MSQA~\cite{linghu2024msr3d}, we only use the subset derived from ScanNet~\cite{dai2017scannet} and follow~\cite{thai2025splattalk} by replacing the interleaved images in situations and questions with text. For VSI-Bench~\cite{yang2024thinkinginspace}, we incorporate the official training split released by~\cite{yang2024thinkinginspace} and the custom training data created by VLM-3R~\cite{fan2025vlm3rvisionlanguagemodelsaugmented}. In order to maintain a balanced contribution across datasets, we subsample approximately half of the available training samples from MSQA~\cite{linghu2024msr3d} and the VSI-Bench data created by VLM-3R~\cite{fan2025vlm3rvisionlanguagemodelsaugmented}.

\begin{table}[h]
    \centering
    \scriptsize
    \caption{\textbf{Detailed training data statistics.}}
    \vspace{-2mm}
    \resizebox{\linewidth}{!}{
    \begin{tabular}{l cccc}
    \toprule
    Source & \# samples & Datasets & \begin{tabular}[c]{@{}c@{}}Question\\ Length\end{tabular} & \begin{tabular}[c]{@{}c@{}}Answer\\ Length\end{tabular} \\
    \midrule
    ScanQA~\cite{azuma2022scanqa} & 26,515 & ScanNet & 8.7 & 2.4 \\
    SQA3D~\cite{ma2022sqa3d} & 79,445 & ScanNet & 28.0 & 1.1 \\
    MSQA~\cite{linghu2024msr3d} & 49,747 & ScanNet & 33.8 & 8.5 \\
    VSI-Bench~\cite{yang2024thinkinginspace} & 9,969 & ScanNet, ScanNet++, ARKitScenes & 39.4 & 1.0 \\
    VSI-Bench (VLM-3R)~\cite{fan2025vlm3rvisionlanguagemodelsaugmented} & 106,890 & ScanNet, ScanNet++, ARKitScenes & 47.9 & 1.0 \\
    \bottomrule
    \end{tabular}}
    \label{tab:dataset}
\end{table}

\paragraph{Evaluation Setup.}
For MSQA~\cite{linghu2024msr3d} and Beacon3D~\cite{huang2025beacon3d}, we adopt GPT-based scoring~\cite{OpenEQA2023,liu2024mmbench} for computing the evaluation metrics. Following the protocol by the respective works, we employ \texttt{gpt-4o-2024-08-06} as the evaluator model and use the evaluation prompt provided in~\cite{OpenEQA2023}.

\section{Robustness to 3D Foundation Model Choice}\label{sec:vggt}
To assess the robustness of our framework to the choice of the underlying 3D foundation model, we train a variant of Loc3R-VLM in which we replace CUT3R~\cite{wang2025cut3r} with VGGT~\cite{wang2025vggt}. For efficiency, this experiment is conducted using a representative subset of the full training data. As shown in~\cref{tbl:vggt}, the VGGT-based variant achieves performance comparable to our primary model. This result demonstrates that our framework is not tightly coupled to CUT3R-specific representations and can readily integrate alternative 3D foundation models.

\vspace{-2mm}
\begin{table}[htbp]
    \centering
    \caption{\textbf{Evaluation with different 3D Foundation Model.} Replacing our default CUT3R encoder with VGGT yields similar performance, demonstrating that our framework is not dependent on a specific 3D representation backbone.}
    \vspace{-2mm}
    \resizebox{0.65\linewidth}{!}{
    \begin{tabular}{lcccc}
    \toprule
    \multirow{2}{*}{Method} & 
    VSI-Bench & 
    \multicolumn{3}{c}{SQA3D} \\
    \cmidrule(lr){2-2} \cmidrule(lr){3-5}
     & Avg. & EM-R & Acc@1m & Acc@30$^\circ$ \\
    \midrule
    Ours w/ CUT3R & 53.0 & 63.5 & 74.2 & 57.1 \\
    Ours w/ VGGT & 54.2 & 63.9 & 74.4 & 59.4 \\
    \bottomrule
    \end{tabular}
    }
    \label{tbl:vggt}
\end{table}

\vspace{-5mm}
\section{Inference Efficiency}\label{sec:efficiency}
In this section, we compare the inference efficiency of our full model (with the CUT3R encoder) against the LLaVA-Video-7B~\cite{zhang2024llava} base model on VSI-Bench~\cite{yang2024thinkinginspace}. We report inference latency and peak memory usage for 32 input frames on a single NVIDIA RTX 4090 (fp16) in~\cref{tbl:latency}. For the VLM, we measure the time-to-first-token (TTFT). The CUT3R encoder introduces only modest memory overhead (+6.8\%), and the total latency of 2.6s for our full model remains well within practical bounds for VLM applications. Importantly, the 3D encoding step is performed only once per video. The resulting CUT3R tokens can be cached and reused across multiple questions about the same video, incurring no additional encoding latency after the initial computation. Consequently, the relative contribution of the 3D encoding cost diminishes as more queries are issued for the same video or as response length increases. During training, CUT3R tokens are pre-computed offline and thus incur no additional runtime overhead.

\begin{table}[h]
    \centering
    \caption{\textbf{Efficiency Analysis.} Inference latency (TTFT) and peak VRAM usage for 32 input frames on a single RTX 4090 (fp16). The CUT3R encoder introduces a one-time per-video encoding cost of 1.2s and a modest increase of +6.8\% in peak VRAM. Once computed, CUT3R tokens can be cached and reused across multiple queries for the same video without incurring additional encoding latency.}
    \vspace{-2mm}
    \resizebox{0.65\linewidth}{!}{
    \begin{tabular}{lccc}
    \toprule
    \multirow{2}{*}{Method} &
    \multicolumn{2}{c}{Latency (s)} &
    \multirow{2}{*}{Peak VRAM (GB)} \\
    \cmidrule(lr){2-3}
    & CUT3R Enc. & Total &  \\
    \midrule
    LLaVA-Video-7B & -- & 1.3 & 19.0  \\
    Ours & 1.2 & 2.6 & 20.3  \\
    \bottomrule
    \end{tabular}
    }
    \label{tbl:latency}
\end{table}

\section{Correlating Localization Accuracy with QA Performance and Uncertainty}\label{sec:effects_loc}
As described in Main Paper Section~3.3, the position head of our situation modeling module outputs both a position estimate $\hat{\mathbf{p}} = [\hat{x}, \hat{y}]^\top$ and a corresponding uncertainty vector $\boldsymbol{\sigma}_{\text{pos}} = [\sigma_x, \sigma_y]^\top$, learned through the Gaussian negative log-likelihood objective. This probabilistic formulation allows the model not only to estimate the agent’s location but also to quantify its confidence in the prediction. To better understand how the internal situation representation of \method{} affects downstream reasoning, we analyze how localization accuracy relates to (i) situated QA performance and (ii) the predicted positional uncertainty. For this analysis, we partition samples from the SQA3D~\cite{ma2022sqa3d} test set according to a localization success criterion: a prediction is considered accurate if the estimated pose achieves a position error of $\leq 1.0$m and an orientation error of $\leq 45^\circ$.

\begin{figure}[htbp]
    \centering
    \includegraphics[width=0.75\linewidth]{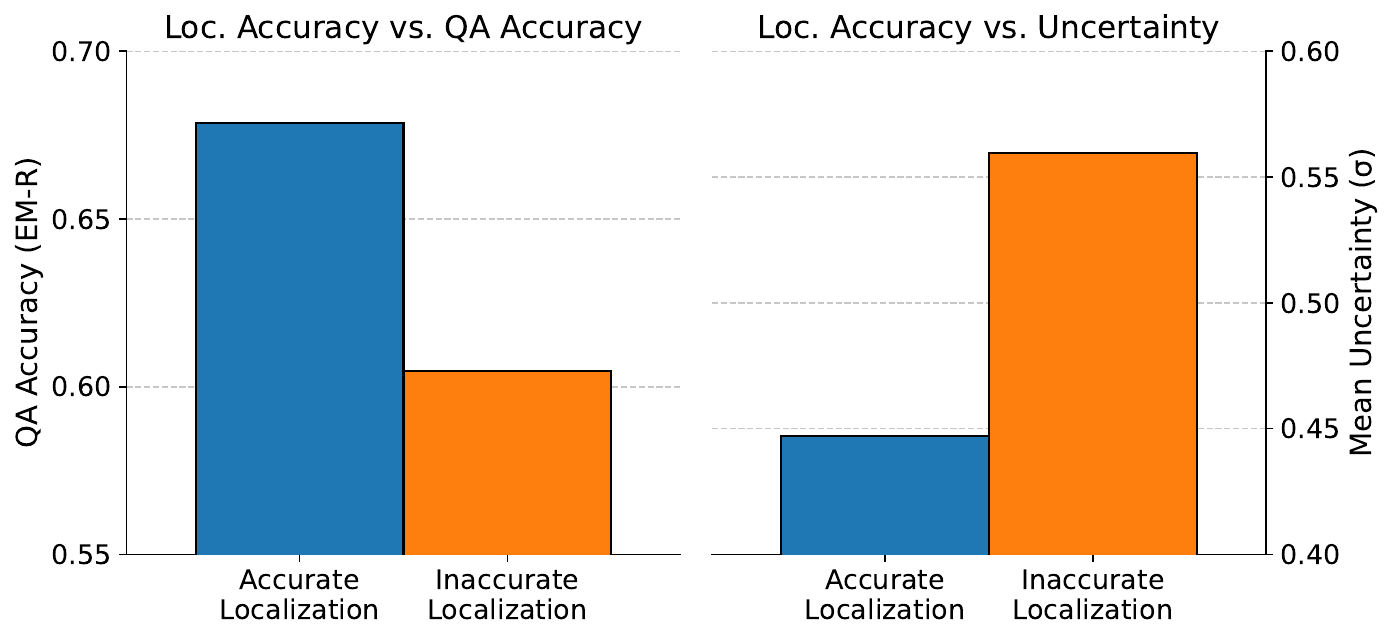}
    \caption{\textbf{Correlation Between Localization Accuracy, QA Performance, and Predicted Uncertainty in \method{}} on SQA3D~\cite{ma2022sqa3d}.
    Localization is considered accurate for a position error $\leq 1.0$m and an orientation error $\leq 45^\circ$.
    \textbf{Left}: QA accuracy (EM-R) is substantially higher when \method{} correctly localizes the agent, demonstrating that its learned situation representation enables effective viewpoint-aware reasoning.
    \textbf{Right}: The predicted positional uncertainty $\sigma_{\text{pos}}$ is notably higher for instances where localization is inaccurate, demonstrating that the model's uncertainty estimates can reflect the reliability of the predicted situation.}
    \label{fig:causal}
\end{figure}

The results in~\cref{fig:causal} reveal two key observations. 
First (left), when \method{} successfully localizes the agent within this threshold, it achieves substantially higher QA accuracy (EM-R), whereas QA performance drops markedly when localization fails. This demonstrates that \method{} effectively leverages its internal situation representation to ground questions in the correct viewpoint, enabling strong situated reasoning. Second (right), localization accuracy is strongly reflected in the predicted positional uncertainty $\sigma_{\text{pos}}$. When the model produces inaccurate position estimates, it also predicts higher uncertainty. This indicates that the probabilistic formulation of our position head yields meaningful confidence estimates that reflect the reliability of the inferred situation.

Overall, these results highlight that the reasoning performance of \method{} is tightly coupled with its internal situation modeling: accurate localization supports viewpoint-aware reasoning, while the predicted uncertainty provides a useful indicator of localization reliability.

\section{BEV Representation}\label{sec:bev}
In this section, we describe how we obtain the ground-truth BEV coordinates for supervising the global layout reconstruction objective during training. For each image patch in the video sequence, we apply the following steps:

\paragraph{Lifting patches to 3D.}
Given a depth map $D$ and camera intrinsics $K$, each image patch is associated with a 3D point corresponding to its center pixel $(u,v)$. Using the patch’s median depth $D(u,v)$, we back-project the pixel to the camera coordinate frame, denoting its homogeneous coordinates as:
\begin{equation}
\mathbf{\bar{p}}^{\text{cam}} =
\begin{bmatrix}
D(u,v)\, K^{-1}[u~~v~~1]^\top \\
1
\end{bmatrix}
\end{equation}
Patches exhibiting high depth variance (e.g., on object edges) are deemed unreliable and thus masked out during loss computation.

\paragraph{Transforming to a canonical frame.}
Following the convention of CUT3R~\cite{wang2025cut3r}, we set the first video frame as the canonical reference. Let $T_{w\leftarrow i} \in \mathbb{R}^{4\times4}$ denote the camera-to-world pose of frame $i$. Each lifted point is transformed into this reference frame by applying the transformations 
\begin{equation}
\mathbf{\bar{p}}^{0} = 
T_{0\leftarrow i}\,\mathbf{\bar{p}}^{\text{cam}},
\qquad
T_{0\leftarrow i} = T_{w\leftarrow 0}^{-1} T_{w\leftarrow i}
\end{equation}
We follow the ScanNet camera frame convention, where the camera frame axes are defined as: $x$-axis points right, $y$–axis points down, and $z$–axis points forward.

\paragraph{Gravity alignment.}
To remove pitch and roll, we estimate a rotation matrix $R_{\text{align}}$ that aligns the reference camera’s gravity direction with the vertical axis. Extracting the Cartesian coordinates $\mathbf{p}^{0} \in \mathbb{R}^3$ from the homogeneous vector $\bar{\mathbf{p}}^{0}$, we apply this rotation to yield gravity-aligned points:
\begin{equation}
\mathbf{p}^{\text{lvl}} = R_{\text{align}}\, \mathbf{p}^{0}.
\end{equation}

\paragraph{Projection onto the ground plane.}
Gravity-aligned points $\mathbf{p}^{\text{lvl}} = (x,y,z)$ are projected onto the ground plane by collapsing along the height dimension to obtain the BEV points:
\begin{equation}
\mathbf{p}^{\text{BEV}} = (x, z).
\end{equation}
These 2D ground-plane coordinates form the BEV ground-truth used to supervise our global layout reconstruction module during training.

\section{Layout Reconstruction}\label{sec:layout_acc}
\paragraph{Layout Reconstruction Evaluation.} Global Layout Reconstruction serves as an auxiliary training objective to enhance the model’s understanding of cross-view spatial relationships and global scene structure. We report the reconstruction accuracy of \method{} in this section, while noting that the model is not explicitly optimized to achieve the best possible performance on this task. We evaluate on 67 scenes from the ScanNet test split, none of which were seen during training. For each vision token, we measure the Euclidean distance between the predicted BEV position and the ground-truth BEV coordinates computed as described in~\cref{sec:bev}. We report Mean Error and RMSE (both in meters), as well as Acc@0.25m, Acc@0.5m, and Acc@1.0m, which quantify the fraction of predictions within the respective distance thresholds.

As shown in~\cref{tab:layout_acc}, the model achieves low reconstruction error, with most predictions falling within a few tens of centimeters of the ground-truth. The high Acc@0.5m and Acc@1.0m numbers indicate that the auxiliary layout objective enables the model to learn a metrically consistent global representation of the scene.

\begin{table}[h]
    \centering
    \caption{
    \textbf{Evaluation of Global Layout Reconstruction} on 67 unseen ScanNet~\cite{dai2017scannet} test scenes. \method{} is able to ground image patches with high spatial accuracy in the BEV space.
    }
    \vspace{-2mm}
    \resizebox{0.65\linewidth}{!}{
    \begin{tabular}{ccccc}
    \toprule
    Mean Error (m) & RMSE (m) & Acc@0.25m & Acc@0.5m & Acc@1.0m \\
    \midrule
    0.40 & 0.48 & 40.6 & 74.8 & 94.4 \\
    \bottomrule
    \end{tabular}}
    \label{tab:layout_acc}
\end{table}

\paragraph{Layout Representation.} Our choice to perform layout reconstruction in a 2D BEV representation is inspired by how humans form cognitive maps, organizing spatial information into compact, low-dimensional abstractions~\cite{Bottini2020KnowledgeAR, Tolman1948}. To validate this design choice, we conduct an ablation study in which the model directly predicts patch coordinates in 3D space. Both variants are trained on a representative subset of the original training data.

As shown in~\cref{tab:bev_ablation}, our chosen 2D BEV representation leads to stronger QA performance and provides comparable localization accuracy. These results suggest that the BEV abstraction is not only cognitively motivated but also beneficial for the model, enhancing spatial reasoning while preserving localization fidelity.
\begin{table}[b]
    \centering
    \caption{
    \textbf{Ablation on Layout Representation.} We compare our 2D BEV representation for global layout reconstruction against directly predicting 3D coordinates. The BEV-based model achieves stronger QA performance while maintaining comparable localization accuracy.
    }
    \vspace{-2mm}
    \resizebox{0.65\linewidth}{!}{
    \begin{tabular}{l c c ccc}
    \toprule
    \multirow{2.5}{*}{\makecell{Layout\\Representation}} &
    \multirow{1}{*}{ScanQA} &
    \multirow{1}{*}{MSQA} &
    \multicolumn{3}{c}{SQA3D} \\
    \cmidrule(lr){2-2}
    \cmidrule(lr){3-3}  
    \cmidrule(lr){4-6}
     & CIDEr & Overall & EM-R & Acc@1.0m & Acc@30$^\circ$ \\
    \midrule
        3D & 101.1 & 58.3 & 62.0 & \textbf{77.2} & 55.4 \\
        2D BEV (Ours) & \textbf{101.7} & \textbf{58.6} & \textbf{63.5} & 74.2 & \textbf{57.1} \\
    \bottomrule
    \end{tabular}}
\label{tab:bev_ablation}
\end{table}

\section{Frame Count Analysis}\label{sec:frame_sampling}
We analyze the sensitivity of \method{} to the number of input frames in~\cref{tbl:frame_count}. Increasing the frame count from 16 to 32 yields a noticeable improvement in both QA performance and localization accuracy, as the model is provided with a more comprehensive view of the 3D scene. However, performance plateaus when scaling further to 40 frames. Overall, the model maintains high stability across the 16--40 frame range, demonstrating its robustness to the density of visual inputs.

\begin{table}[h]
    \centering
    \caption{\textbf{Evaluation for Varying Input Frame Count.} The model maintains stability across different numbers of input frames.}
    \vspace{-2mm}
    \resizebox{0.6\linewidth}{!}{
    \begin{tabular}{lcccc}
    \toprule
    \multirow{2}{*}{Frame Count} &  
    VSI-Bench & 
    \multicolumn{3}{c}{SQA3D} \\
    \cmidrule(lr){2-2} \cmidrule(lr){3-5}
     & Avg. & EM-R & Acc@1m & Acc@30$^\circ$ \\
    \midrule
    16 & 60.3 & 64.1 & 70.7 & 59.8 \\
    32 & 63.2 & 65.0 & 75.9 & 63.0 \\
    40 & 63.3 & 64.8 & 75.5 & 62.6 \\
    \bottomrule
    \end{tabular}
    }
    \label{tbl:frame_count}
\end{table}

\section{Qualitative Results}\label{sec:supp_qualitative}
\begin{figure*}[!t]
    \centering

    \includegraphics[width=\linewidth]{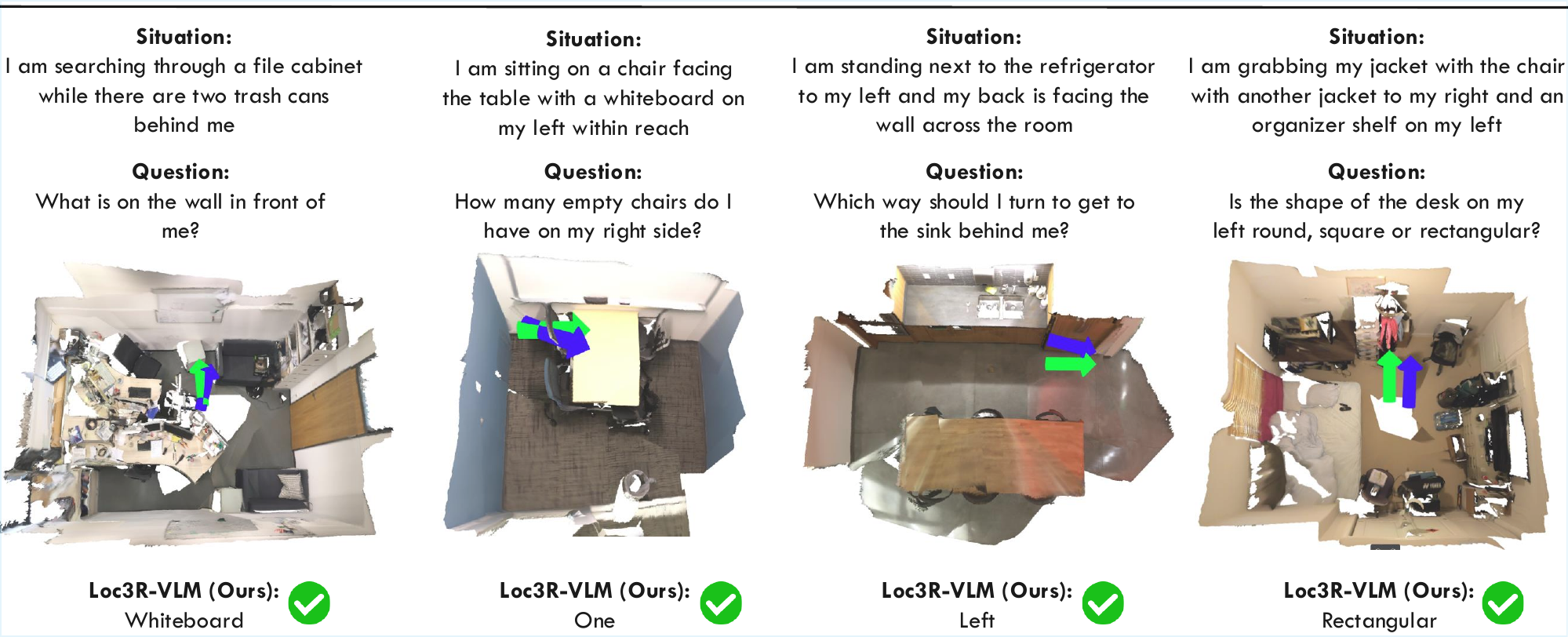}\vspace{1mm}

    \includegraphics[width=\linewidth]{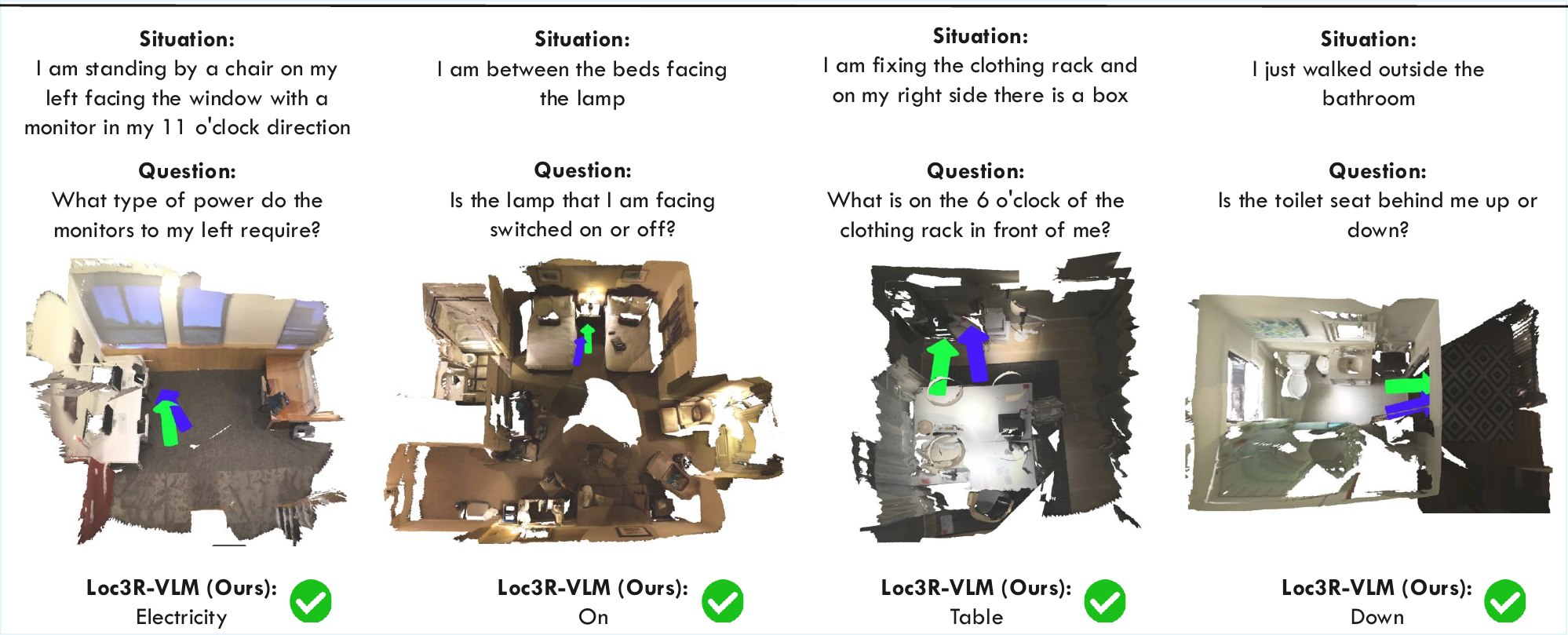}\vspace{1mm}

    \includegraphics[width=\linewidth]{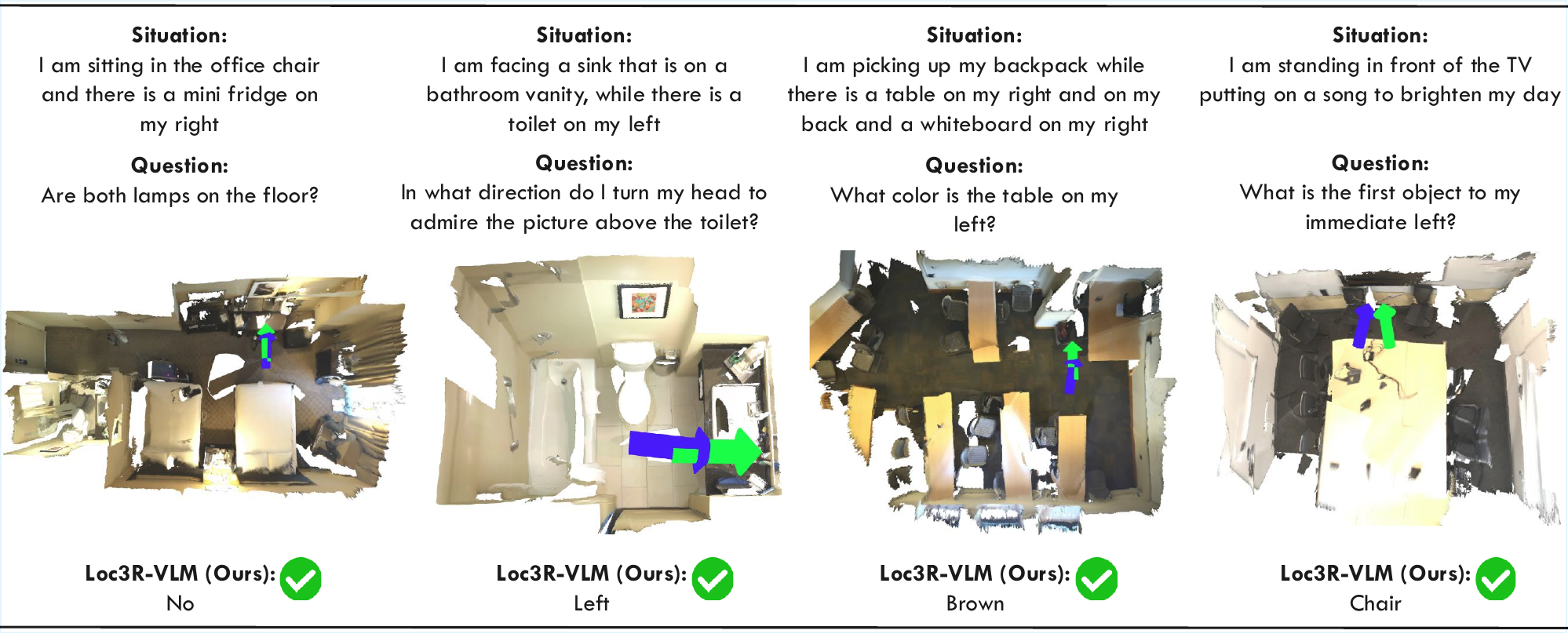}\vspace{1mm}

    \caption{
        \textbf{Qualitative Examples} of language-based localization and situated question answering on SQA3D~\cite{ma2022sqa3d} (\textcolor{blue}{blue}: predicted situation, \textcolor{green}{green}: ground-truth situation).
    }
    \label{fig:additional_qualitative1}
\end{figure*}

\begin{figure*}[!t]
    \centering

    \includegraphics[width=\linewidth]{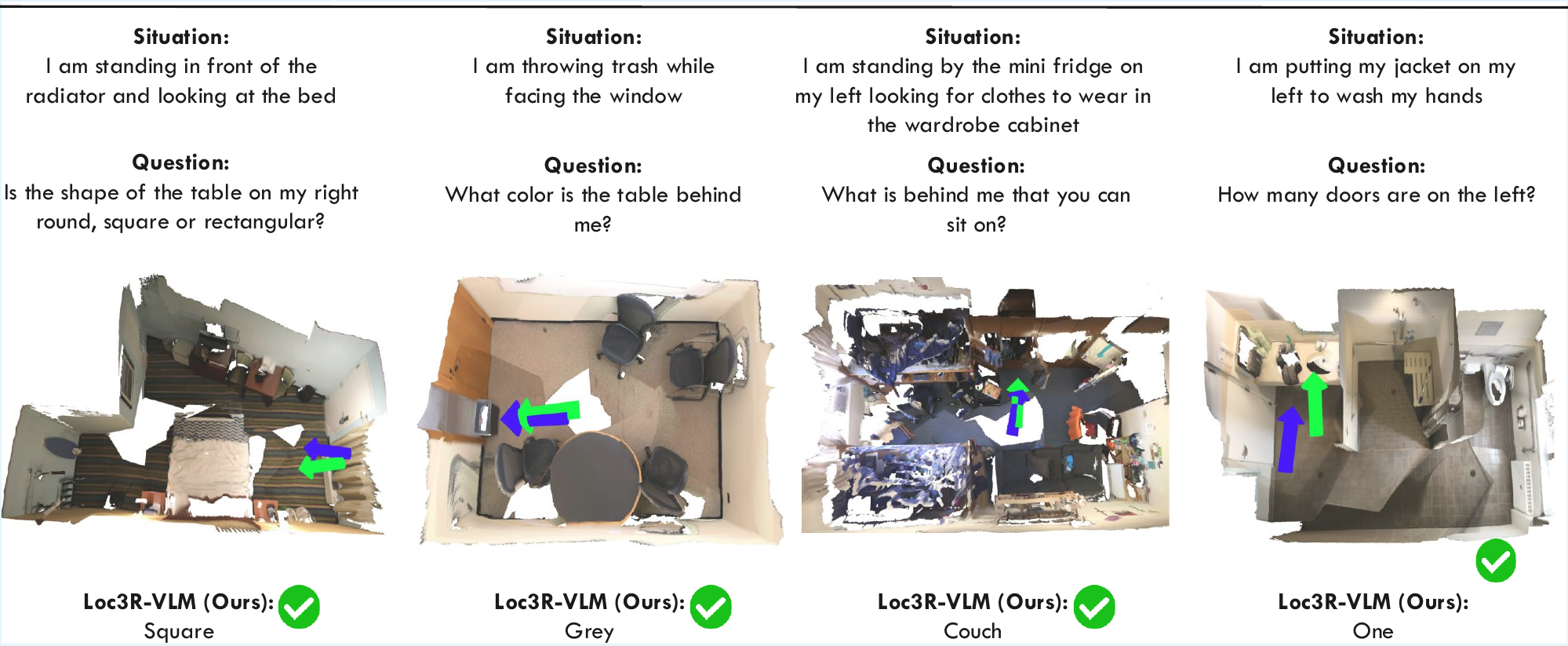}\vspace{1mm}

    \includegraphics[width=\linewidth]{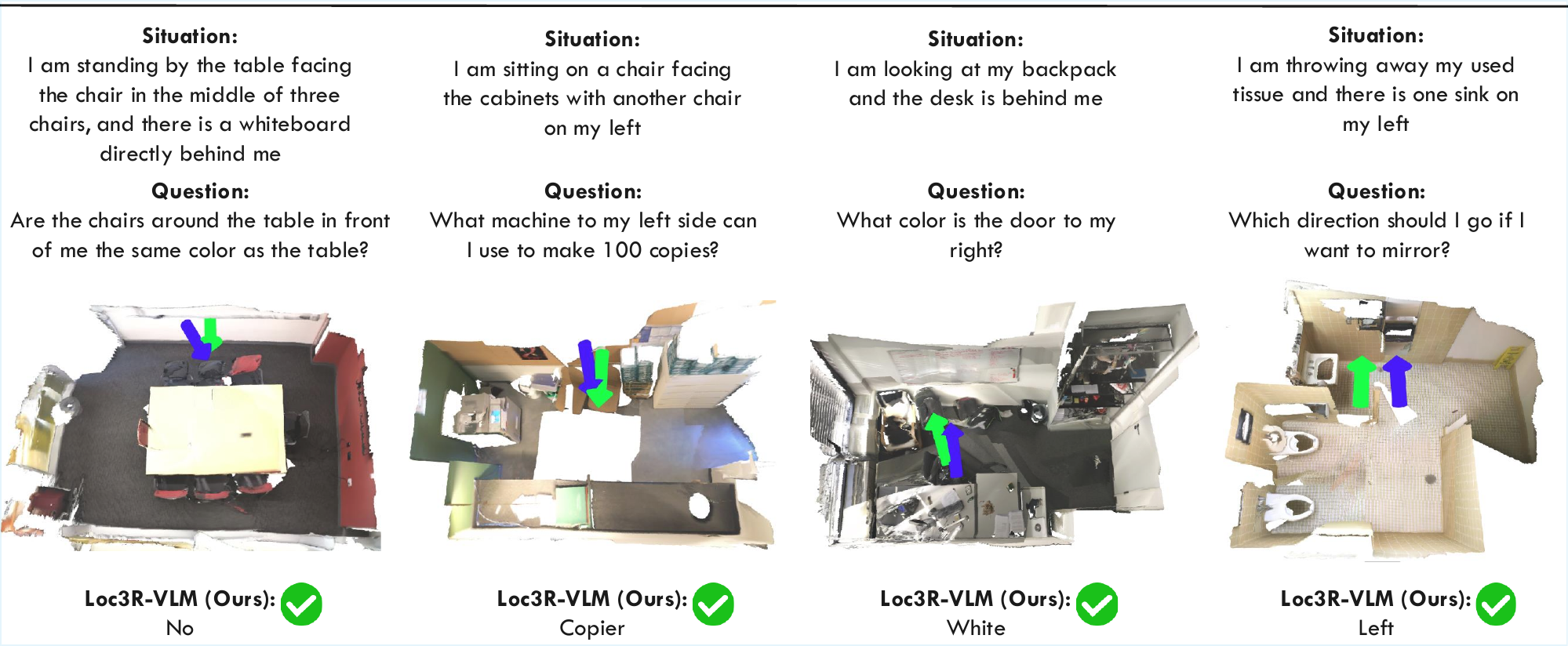}\vspace{1mm}

    \includegraphics[width=\linewidth]{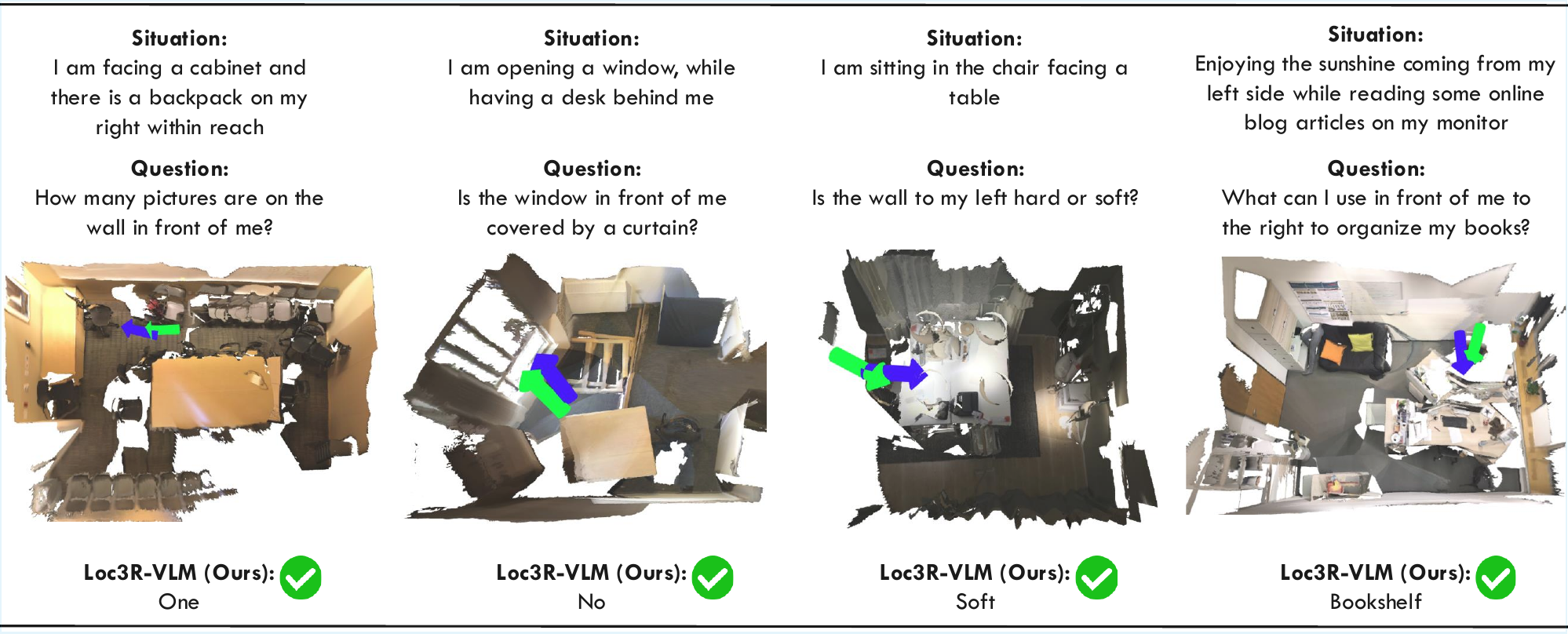}\vspace{1mm}

    \caption{
        \textbf{Additional Qualitative Examples} of localization and situated question answering on SQA3D~\cite{ma2022sqa3d} (\textcolor{blue}{blue}: predicted situation, \textcolor{green}{green}: ground-truth situation).
    }
    \label{fig:additional_qualitative2}
\end{figure*}

In~\cref{fig:additional_qualitative1} and~\cref{fig:additional_qualitative2}, we illustrate more qualitative examples of \method{} for language-based localization and situated question answering on SQA3D~\cite{ma2022sqa3d}. \method{} demonstrates strong situational awareness and spatial understanding across diverse scenes and situation-question pairs.

\section{Failure Cases}
\label{sec:supp_failures}
\begin{figure*}[!t]
    \centering

    \includegraphics[width=\linewidth]{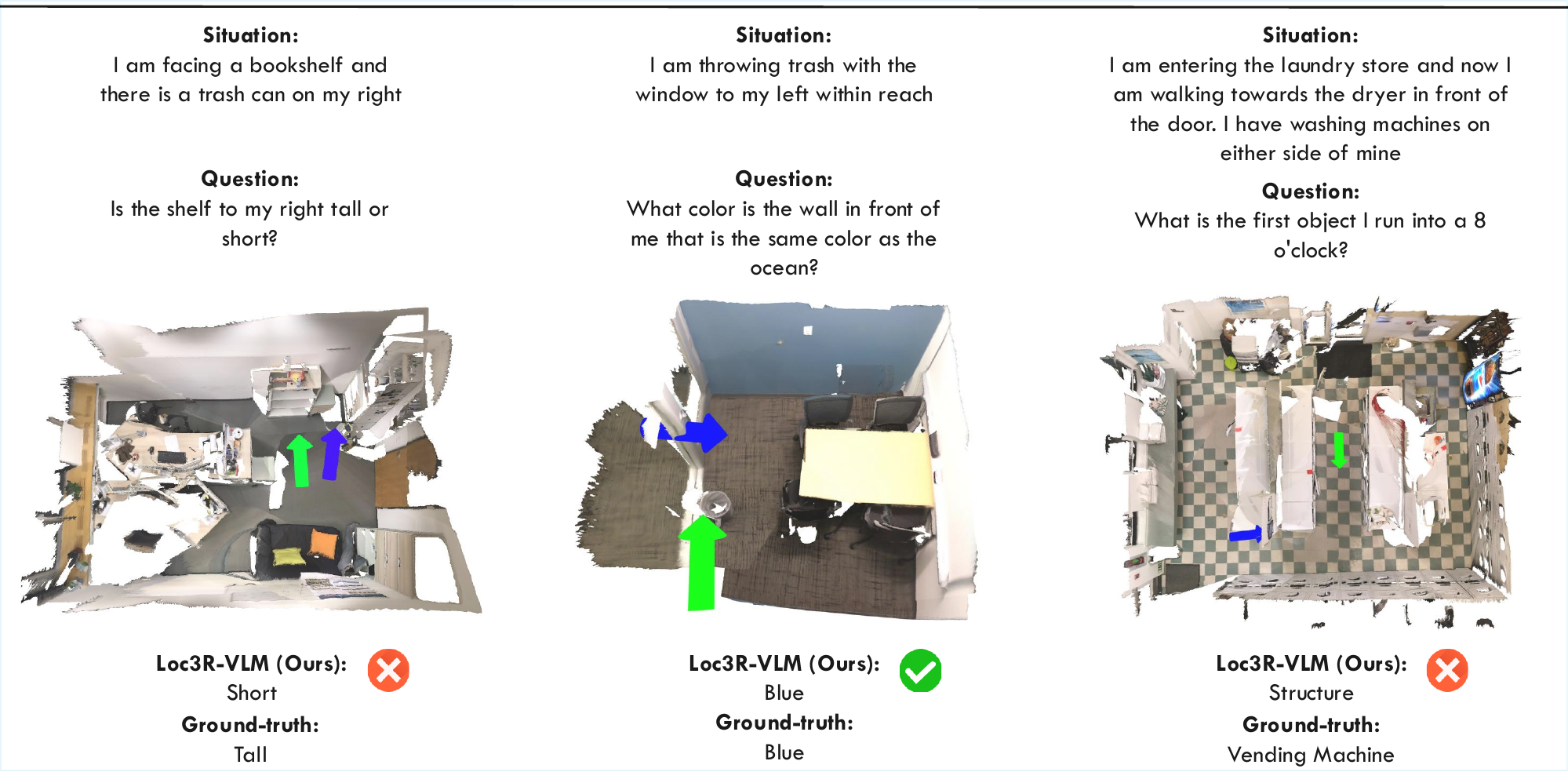}

    \includegraphics[width=\linewidth]{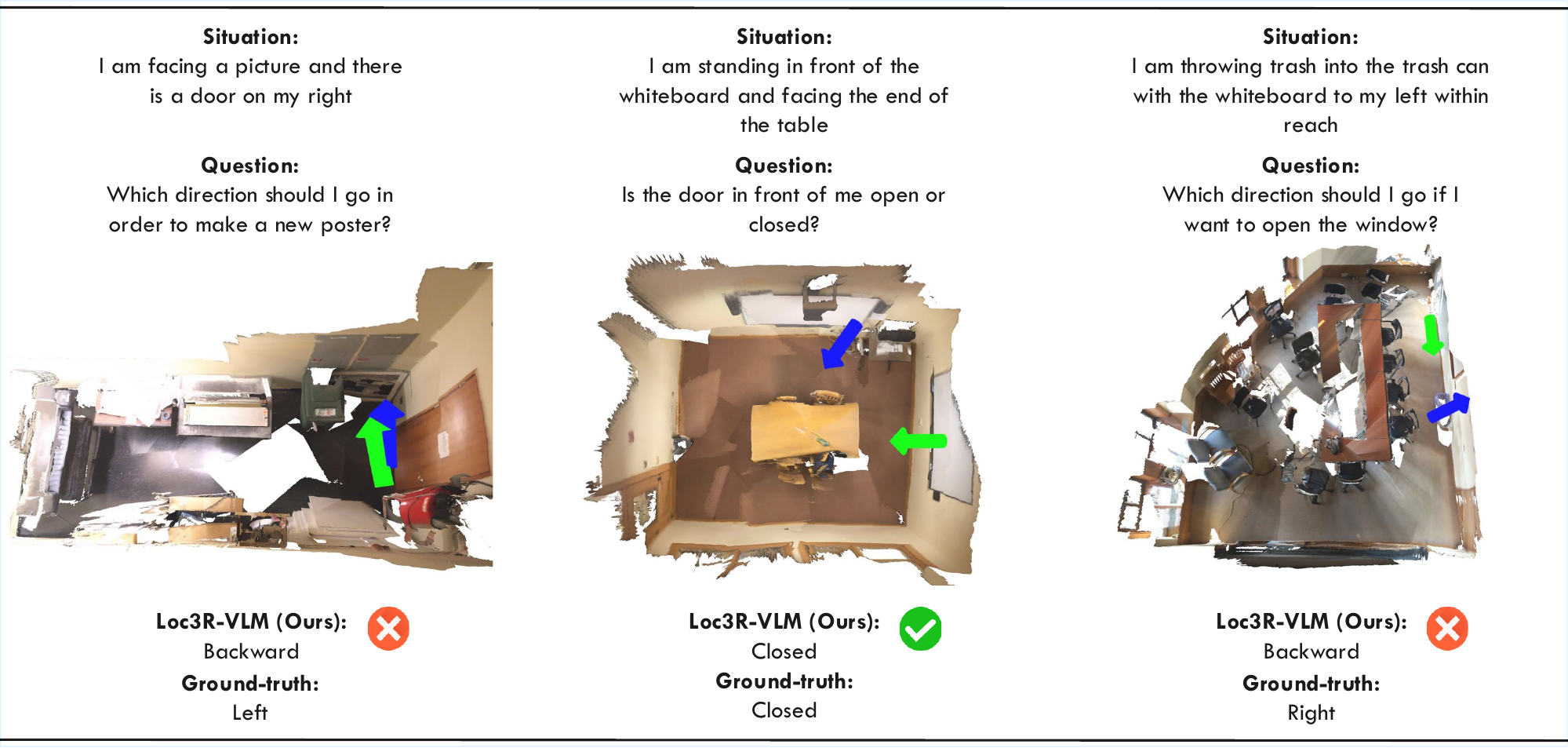}\vspace{1mm}

    \caption{
        \textbf{Visualization of Failure Cases.} We show three different types of failure cases. Left: correct localization but wrong answer. Middle: wrong localization but correct answer. Right: wrong localization and wrong answer (\textcolor{blue}{blue}: predicted situation, \textcolor{green}{green}: ground-truth situation).
    }
    \label{fig:failure_cases}
\end{figure*}

In~\cref{fig:failure_cases}, we visualize representative failure modes of our model for localization and situated reasoning, grouped into three categories:
(1) \textbf{Correct localization, wrong QA (left).} The model grounds the described situation correctly but outputs an incorrect answer. These failures indicate that errors can still arise in the downstream reasoning stage even when the situation is accurately inferred.
(2) \textbf{Incorrect localization, correct QA (middle).} Here, the predicted situation is inaccurate, yet the model answers correctly. As illustrated in the examples, this often occurs when the localization is challenging, while the question itself is easy to answer -- either through common knowledge or because the referenced object appears only once.
(3) \textbf{Incorrect localization and incorrect QA (right).} These failures typically arise in complex scenes or under ambiguous situation descriptions, where the model struggles to accurately infer the agent’s pose. In the examples, the predicted answer is nevertheless consistent with the incorrectly inferred viewpoint. This indicates that the reasoning process respects the internally predicted situation representation. Consequently, localization errors can propagate to the reasoning stage, leading to incorrect answers despite otherwise plausible reasoning.

\section{Limitations and Future Work}\label{sec:supp_limitations}
Despite its strong performance, \method{} exhibits several limitations:

\paragraph{Vertical Granularity.} By projecting 3D information into a 2D bird's-eye-view (BEV) representation, our framework inherently discards vertical detail. This can hinder reasoning in multi-floor environments or for tasks requiring precise height-based distinctions, such as identifying objects on vertically stacked surfaces. Future research could explore layered BEV architectures or the integration of object-centric tokens to reintroduce the vertical dimension while maintaining computational efficiency.

\paragraph{Scene Coverage.} The construction of the global cognitive map is currently constrained by the fixed-length frame sampling (i.e., 32 frames). In expansive scenes with low viewpoint overlap, this sparse sampling may result in ``blind spots'' that can propagate into downstream localization or reasoning errors. To address this, future work could investigate spatially adaptive or coverage-aware frame selection strategies to ensure more comprehensive scene observation without exceeding the model's context window.

\paragraph{Domain Scope.} Our approach focuses exclusively on static indoor scenes. Adapting our framework to handle dynamic scenes and outdoor environments remains an interesting direction for future work.

\clearpage

\end{document}